\pdfoutput=1

\documentclass[11pt]{article}

\usepackage[final]{acl}

\usepackage{blindtext}
\usepackage{hyperref}
\usepackage{enumitem}

\usepackage{times}
\usepackage{latexsym}

\usepackage[T1]{fontenc}

\usepackage[utf8]{inputenc}

\usepackage{microtype}

\usepackage{inconsolata}

\usepackage{graphicx}
\usepackage{booktabs}
\usepackage{mathtools} 
\usepackage{amsmath}
\usepackage{xcolor}

\title{LLMs vs Established Text Augmentation Techniques for Classification: When do the Benefits Outweight the Costs?}

\author{Jan Cegin$^{\spadesuit}$$^\dagger$, Jakub Simko$^\dagger$,
{\bf Peter Brusilovsky}$^\ddagger$ \\
  $^{\spadesuit}$ Faculty of Information Technology, Brno University of Technology, Brno, Czechia \\
  $^\dagger$ Kempelen Institute of Intelligent Technologies, Bratislava, Slovakia\\
  $^\ddagger$ University of Pittsburgh, Pittsburgh, USA \\
    \texttt{\{jan.cegin jakub.simko\}}@kinit.sk, \texttt{peterb@pitt.edu }}

\begin{document}
\maketitle
\begin{abstract}
The generative large language models (LLMs) are increasingly being used for data augmentation tasks, where text samples are LLM-paraphrased and then used for classifier fine-tuning. However, a research that would confirm a clear cost-benefit advantage of LLMs over more established augmentation methods is largely missing. To study if (and when) is the LLM-based augmentation advantageous, we compared the effects of recent LLM augmentation methods with established ones on 6 datasets, 3 classifiers and 2 fine-tuning methods. We also varied the number of seeds and collected samples to better explore the downstream model accuracy space. Finally, we performed a cost-benefit analysis and show that LLM-based methods are worthy of deployment only when very small number of seeds is used. Moreover, in many cases, established methods lead to similar or better model accuracies.
\end{abstract}

\section{Introduction}

The emergence of recent large language models (LLMs) such as GPT-4, Gemini, Llama, and their wide availability, prompted their use in \emph{augmentation} of textual datasets~\cite{ubani2023zeroshotdataaug, dai2023auggpt, piedboeuf-langlais-2023-chatgpt, cegin-etal-2023-chatgpt, cegin2024effectsdiversityincentivessample}. In most LLM-based augmentation scenarios, the dataset size is increased through paraphrasing of original samplesThe extended datasets are then used for training small \emph{downstream} classifiers with small inference costs. LLM augmentation has been used in various domains such as sentiment analysis~\cite{ONAN2023101611, piedboeuf-langlais-2023-chatgpt}, intent classification~\cite{cegin-etal-2023-chatgpt}, news classification~\cite{piedboeuf-langlais-2023-chatgpt, cegin2024effectsdiversityincentivessample} and health symptoms classification~\cite{dai2023auggpt}.

While LLM augmentation improves downstream classifiers, it is also costly (power consumption, CO$_2$ emissions), as generative models often feature parameters in tens of billions. This is magnitudes higher than other \emph{established} (most used) augmentation methods, including \emph{back translation} paraphrasing, or BERT-based \emph{word insertion} and \emph{synonym swap}. A comparison of established methods with newer LLM-based methods could explore the cases where the established (and much cheaper) methods are preferable due to their equal/better downstream classifier performance (e.g., accuracy) or better cost-benefit ratio.
In this line, previous works~\cite{piedboeuf-langlais-2023-chatgpt, ubani2023zeroshotdataaug, dai2023auggpt} have measured classifier performance, comparing \emph{LLM-based} and \emph{established} augmentation methods. The results have so far been conflicting and mixed. Furthermore, existing studies were limited in terms of parameters: neglecting the variety of available LLMs, the potential impact of the number of seed samples and collected samples, and the variety of classifiers and their fine-tuning methods. 

The goal of this paper is to compare the accuracy and cost-benefits of the most used \emph{established} text augmentation methods with their recent \emph{LLM-based} counterparts. Compared with previous studies, this paper offers a more systematic and finer-grained comparison over multiple dimensions. We formulate the following research questions:

\begin{description}[labelwidth = 24pt, leftmargin = !]
    \item[RQ1:] \emph{Considering downstream classifier accuracy, in which cases do the established textual augmentation methods work equally or better than the LLM-based methods?}
    \item[RQ2:] \emph{In which cases does the cost of using LLM-based textual augmentation methods instead of established ones outweigh its benefits?}
\end{description}

We empirically investigated three techniques commonly used in textual augmentation: \emph{paraphrasing}, \emph{word inserts} and \emph{word swaps} (replacements). All three exist in both \emph{established} and \emph{LLM-based} variants. In the established variant, paraphrasing is done through back-translation using a RNN~\cite{sennrich-etal-2016-improving}, while inserts and swaps use BERT-based approach~\cite{kobayashi-2018-contextual, kumar-etal-2020-data}. For LLM-based variants, we prompted 2 LLMs (GPT-3.5 and Llama-3\footnote{Albeit BERT is often referred to as an early LLM, for the sake of wording clarity, we do not consider it as such in our study.}) to perform all three techniques. We experimented with 6 different datasets (with tasks of sentiment analysis, news classification, and intent classification), 3 downstream classifier models (BERT, RoBERTa, DistilBERT), and 2 fine-tuning approaches (fully fine-tuned, and QLoRA~\cite{dettmers2024qlora}). Furthermore, we investigated various numbers of seed and collected samples used in the augmentation. Together, this resulted in a total of 267,300 fine-tunings, from which we identified the best performing LLM and established methods (answering Q1). These were then further scrutinized under cost-benefit analysis (answering Q2).

The most prominent findings are: 1) The best LLM augmentation methods outperform established ones  \emph{only} when a small number of seeds is used. The advantage of LLM-based augmentation diminishes with increased seed numbers, making it less cost-feasible. This hints towards using LLM-based methods only in scenarios with a small number of seeds per label (5-20).
2) LLM augmentation methods have higher impact on accuracy of less robustly pre-trained classifiers such as DistilBERT or BERT.
3) LLM augmentation methods have higher impact on classifier accuracy for full fine-tuning when compared to QLoRA fine-tuning.

\section{Related Work: Text Augmentation}

Text augmentation is a process of increasing the diversity of training text data without necessarily collecting more original (or seed) data. Text augmentation was inspired by image augmentation~\cite{feng-etal-2021-survey, zhou2024surveydataaugmentationlarge} where various techniques such as cropping, rotating, flipping, etc. were used to build models that are more robust to image variation and in turn enhance their performance. Text and data augmentation have an increasing number of Google weekly trend searchers in recent years~\cite{feng-etal-2021-survey, zhou2024surveydataaugmentationlarge}, indicating an increasing interest in these kinds of model performance enhancing methods. 

One of the most established are character-based augmentations~\cite{wei-zou-2019-eda, karimi-etal-2021-aeda-easier}, where given a seed text, a new sample is created via character insertion, replacement, or deletion. Another method is backtranslation~\cite{sennrich-etal-2016-improving}, which translates a given text into one language to then translate it back, essentially creating a paraphrase. Various LLMs such as GPT-2~\cite{radford2019language} or BART~\cite{lewis-etal-2020-bart} have also previously been used to create paraphrases. Additional extensions used style transfer to create paraphrases of a certain linguistic style~\cite{Krishna2020}, syntax control of the generated paraphrases~\cite{goyal-durrett-2020-neural, chen-etal-2020-semantically}, multi-lingual paraphrases~\cite{thompson-post-2020-paraphrase} and LLM fine-tuning using QLoRA for specific domains~\cite{chowdhury2022novelty}. Another established method is the usage of pre-trained LLMs to generate new samples by either word insertion or replacement of words via masking certain parts of the seed text and allowing the LLM to find good replacements for the masked parts of the text~\cite{kobayashi-2018-contextual, kumar-etal-2020-data}. 

Text augmentation methods were adapted with the rise of new LLMs such as GPT-4 or Llama to leverage these new powerful models to generally create paraphrases of given seed texts. A recent study~\cite{piedboeuf-langlais-2023-chatgpt} found that while the GPT-3.5 paraphrasing provides an increase in classifier accuracy, it does not outperform the previous established text augmentation methods to a significant margin. In contrast, two studies reported better performance in using LLMs as data augmenters than using previous state-of-art techniques in both the paraphrasing of existing texts~\cite{dai2023auggpt} and in a zero-shot setup of generating new texts using specific prompts~\cite{ubani2023zeroshotdataaug}. Regardless of the mixed results reported, newer LLMs have been used for a variety of augmentation tasks and domains such as automated scoring~\cite{fang2023using}, low-resource language generation~\cite{ghosh-etal-2023-dale}, sentiment analysis ~\cite{piedboeuf-langlais-2023-chatgpt, ubani2023zeroshotdataaug, ONAN2023101611}, news classification~\cite{piedboeuf-langlais-2023-chatgpt}, content recommendation~\cite{contect-based-recom} and health symptoms classifications~\cite{dai2023auggpt}.

Given the wide usage of LLM augmentation methods and the mixed results of studies~\cite{dai2023auggpt, piedboeuf-langlais-2023-chatgpt, ubani2023zeroshotdataaug} comparing them with established augmentation methods, a finer analysis of cases where one of these types of methods is preferable is required. 

\begin{figure*}[t]
    \centering
    \includegraphics[width=14cm]{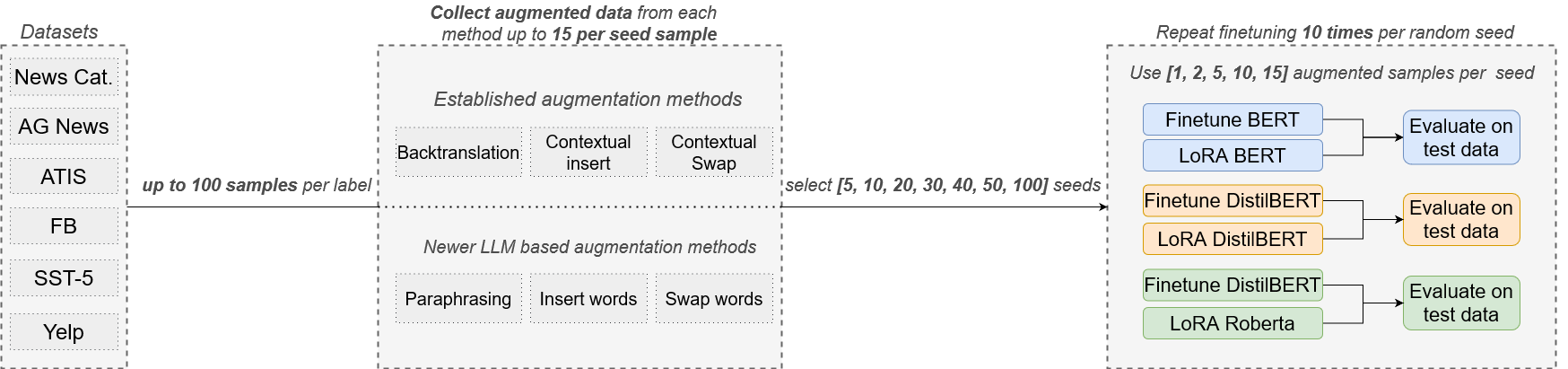}
    \caption{Overview of our methodology. For each dataset, we randomly sample 100 samples per label which are then used to collect up to 15 augmented samples per each seed samples. These seeds are then randomly sampled in various sizes and used for fine-tuning with various number of augmented samples to evaluate each method.}
    \label{fig:dataset_build}
\end{figure*}

\section{Study Design}
\label{sec:data_coll}

To assess advantages of either \emph{established} or newer \emph{LLM-based} augmentation methods, we performed a comparative study. At its core was the same basic scenario (see also figure~\ref{fig:dataset_build}): on a given text classification dataset, a number of seed samples was selected for each class. For each seed sample, a given augmentation method generated a number of additional ``augmented'' samples. Both original data and the augmented samples were then used for fine-tuning of a downstream classifier. This scenario was repeated for all examined methods and a variety of parameters (see below), resulting in a total of 37,125 augmented samples and 267,300 fine-tunings. Then, accuracy of the resulting classifiers was compared to answer Q1. To answer Q2, the augmentation costs in terms of computation time, finances, and CO$_2$ emissions were determined and weighted against accuracy gains in a cost-benefit analysis. We publish all of our measurements, the code and data used~\footnote{Data and code at: \url{https://github.com/kinit-sk/llms_vs_nlpaug_data_aug}}.

The study had following parameters:
\begin{itemize}
    \setlength\itemsep{0em}
    \item the augmentation technique (paraphrasing, contextual word insert, word swap -- realized using either established or LLM-based methods),
    \item the number of seed samples per label (5, 10, 20, 30, 40, 50, 100)\footnote{For some datasets (ATIS, FB) the maximum seed number was lower than 100 due to smaller class sizes, for more details see Appendix~\ref{sec:appendix_dataset_details}},
    \item the number of collected samples per seed (1, 2, 5, 10, 15), 
    \item the LLMs used as augmenters in case of LLM-based methods (GPT-3.5, Llama-3-8B),
    \item the fine-tuned classifiers (DistilBERT, RoBERTa, BERT),
    \item the fine-tuning approach (full, QLoRA),
    \item and the dataset/task (6 datasets). 
\end{itemize}

\subsection{Established Text Augmentation Methods}
\label{sec:established}

As the \textit{established} text augmentation methods, we chose 3 well-known, yet simple and relatively efficient methods as shown by a previous study~\cite{dai2023auggpt}. We went with model-based techniques that leverage some form of trained Seq2Seq model or contextual embedding methods that use smaller LLMs (BERT is frequently considered an early LLM). First among them is the \textit{backtranslation}~\cite{sennrich-etal-2016-improving} -- in the past (before the advent of LLMs) a popular method used for paraphrasing. The method translates a sentence from one language to another and back to create paraphrases. 
Another popular and relatively simple method is the replacing or inserting of words based on embeddings. In our experiments, we used contextual embeddings~\cite{kobayashi-2018-contextual, kumar-etal-2020-data}. We used two contextual embedding methods: \emph{contextual word insertion} and \emph{contextual word swap} (replacement). As a first step, the \emph{contextual word insertion} method randomly inserts masks between words in a sentence, while the \emph{contextual word swap} method randomly replaces a set number of words in the sentence for masks. Next, a model is queried to get the most likely tokens for each mask. The details of these methods and the parameters used for each of these methods can be found in Appendix~\ref{sec:appendix_established_params_used}. We used the implementations provided by the NLPAuglibrary~\cite{ma2019nlpaug}.

\subsection{LLM-based Text Augmentation Methods}
\label{sec:new}

As the \textit{LLM-based} text augmentation methods, we implemented the three given techniques using prompts similar to previous works~\cite{cegin-etal-2023-chatgpt, piedboeuf-langlais-2023-chatgpt}. However, as the previous works generally used only paraphrasing, we devised new prompts explicitly asking the model to replace words for their synonyms or change the text by inserting words to it. Thus, we had 3 different LLM-based text augmentation methods: \emph{paraphrasing} where we asked the model to produce a paraphrase, \emph{word insertion} where we asked the model to produce a new sample by inserting words into seed sample, and \emph{word swap} where we asked the model to produce a new sample by replacing words for their synonyms in the seed sample. We did not specify how much the sentences should be changed to keep the prompts as simple as possible. We used these 3 methods to gather data using both GPT-3.5 and Llama-3-8B. Further details about prompt templates, model types used and parameters used during inference can be found in Appendix~\ref{sec:appendix_llm_params_used}.

\subsection{Datasets}
\label{sec:datasets}

To explore diversity of augmentation effects, we used 6 different datasets, representing three distinct text tasks: the classification of sentiment, of intent, and of news domains. All datasets were multi-class and English. We used  the \emph{News Category}~\cite{misra2022news, misra2021sculpting} and \emph{AG news}~\cite{zhang2015character} for news classification, \emph{FB}~\cite{schuster-etal-2019-cross-lingual} and \emph{ATIS}~\cite{hemphill-etal-1990-atis} for intent classification, and \emph{SST-5}~\cite{socher-etal-2013-recursive} and \emph{Yelp}~\cite{zhang2015character} for sentiment classification. When measuring accuracy of downstream classifiers, we used test splits of each of these datasets. To achieve uniform sizes and distributions, we selected a subset of classes and down-sampled some of them for the use in our experiments. Details about the datasets, labels and class sizes used for each dataset can be found in Appendix~\ref{sec:appendix_dataset_details}.

\subsection{Evaluation Process}

For each combination of number of seeds and datasets, seed samples were randomly selected from among dataset's classes. Then, the selected augmentation method was applied to generate the additional samples. 

We manually checked the \emph{validity} of a random subset (10\%) of the collected data (i.e., whether the created samples truly are paraphrases retaining the labels of their seeds). Previous works have already shown that the validity with newer LLM augmentation methods is high~\cite{cegin-etal-2023-chatgpt, cegin2024effectsdiversityincentivessample}, yet we still sought to confirm it and examine the established methods as well. We found the highest validity of samples for the LLM-based \emph{paraphrasing} with 100\% valid samples. Both LLM-based and established \textit{word insert} and \textit{word swap} methods achieved 95\%-97\% validity, struggling mostly with incorrect named entities. The established paraphrasing \emph{backtranslation} method yields 98\%-99\% valid samples, but also a very large portion of duplicates (around 80\%). Details on validity checks can be found in Appendix~\ref{sec:appendix_para_valid}.

We used BERT-base, DistilBERT-base and RoBERTa-base for fine-tuning. We used the versions of the models from Huggingface and found the best working hyperparameters via hyperparameter search. Hyperparameters with the QLoRA fine-tuning setup can be found in Appendix~\ref{sec:appendix_finetuning_details}. We trained each model 10 times per random seed and used 3 different random seeds with differently sampled seed samples and augmented samples for those seeds to avoid randomness of outcomes. The random seeds ensured that across various combinations, the same seeds and augmented samples could be used. The models were trained separately on the data collected from GPT-3.5 and Llama-3. As we aimed to compare the newer and established augmentation methods in a variety of cases, we used various number of seeds samples per label and number of collected samples per seed sample during fine-tuning. We ended up with a total number of fine-tunings (both full and using QLoRA) at 267,300, as we fine-tuned the models 10 times for each augmentation method, dataset, number of seed samples per label, number of collected samples per seed and random seed combination. 

Finally, we computed the accuracy of all fine-tuned classifiers to allow their comparison.

\section{Study Results}

Our study has multiple parameter dimensions that together yield more than 11 thousand combinations. To keep the result presentation manageable, we collapse some of these dimensions (each of them with a different reasoning).

One dimension we could simplify are the augmentation methods themselves. To keep the comparison of \emph{established} and \emph{LLM-based} methods simple, we only compared best-performing methods from each group (best downstream model accuracy). While the \emph{established} method group contained 3 methods (given by the 3 augmentation techniques and their established implementations), the \emph{LLM-based} method group contained 6 methods (the same 3 techniques, each implemented by 2 different LLMs\footnotemark{}). We performed this comparison for each parameter combination of number of seeds, number of augmented samples per seed, classifier, fine-tuning approach, and dataset. 
\footnotetext{The results from Llama-3 and GPT-3.5 augmentation methods are both labeled as ``LLM methods'', as during the analysis we found no significantly different model accuracy for augmentations created from the two LLMs used for training classifier.}

Among the LLM-based methods, the \emph{paraphrasing} technique performed best in 56\% cases, followed by \emph{word insert} which topping 30\% of cases, and \emph{word swap} with 14\% of cases. Although \emph{paraphrasing} performed best overall, the \emph{word insert} worked best when the RoBERTa classifier was fine-tuned. Among the established methods, the \emph{contextual word insert} performed best in 56\% cases, followed by (backtranslation) \emph{paraphrasing} topping 26\% cases, and \emph{contextual word swap} with 18\% cases. Furthermore, the backtranslation had a stronger effect on classifier accuracy with full fine-tuning and lesser with QLoRA. Given these results, we decided to focus on the comparison of the LLM-based \emph{paraphrasing} with the established \emph{contextual word insert} methods. See appendix~\ref{sec:appendix_combination} for other method comparisons.

Another dimension we could collapse was the \emph{number of collected samples per seed}, where we selected only the most accurate classifier for the same combination of other parameters. However, full details on how the number of collected samples per seed influences the classifier accuracy can be found in Appendix~\ref{sec:appendix_effects_of_no_col_samples}.

\subsection{Classifier Accuracy (RQ1)}\label{sec:established_newer_strict_comp}

To answer the RQ1, we compared the downstream classifier accuracy of LLM-based \emph{paraphrasing} with the established \emph{contextual word insert}, see Table~\ref{tab:count_perf_para_vs_cont_ins}. We counted the number of cases where one of these methods performed statistically significantly better than the other and also the number of cases where there was no statistically significant difference between the two methods, which we denote as the two methods having similar accuracy. For this we used Mann-Whitney-U tests with \emph{p=0.05}.

In most cases, the accuracy of LLM-based \emph{paraphrasing} cannot be statistically distinguished from the \emph{contextual insert}. However, when differences are observed, the LLM-based \emph{paraphrasing} beats the \emph{contextual insert} method in more cases. For full fine-tuning, this can be observed consistently (with the sole exception of RoBERTa with the News Category dataset). For QLoRA, the results are more mixed: while LLM-based \emph{paraphrasing} generally yields better results for BERT and DistilBERT (with exceptions), for RoBERTa the \emph{contextual insert} surpasses the LLM-based \emph{paraphrasing} more often. It should also be noted that of the three classifiers, RoBERTa performed best in \(\sim \)80\% of cases as can be seen in Appendix~\ref{sec:appendix_best_classifier_model}.

An investigation of the difference in mean accuracy between models trained using LLM \textit{paraphrasing} and \textit{contextual insert} can be found in Figure~\ref{fig:mean_diff_perf_para_vs_cont_ins} for various number of seeds per label. Generally, a lower amount of seeds leads to a higher accuracy of classifiers trained on augmented data collected via the LLM \textit{paraphrasing}, than when \textit{contextual insert} is used. This difference in accuracy is highest for 5 to 20 samples per label in cases where \textit{paraphrasing} is more advantageous and decreases with more seed samples used. 

There are also notable cases where \textit{contextual insert} (a far cheaper augmentation method) provides better classifier accuracy than the LLM \textit{paraphrasing}. This can be seen for BERT QLoRA fine-tuning in the FB dataset and Yelp dataset, and RoBERTa fine-tuning for the News Category dataset. BERT QLoRA exhibited results that favor one of the two methods more strongly than other types of fine-tuning and model combinations. In terms of increased classifier accuracy, when comparing fine-tuning with only seed samples themselves, both methods provide a relatively high increase of accuracy, compared to using only seed samples for training classifiers when using QLoRA (see visualization of this in Appendix~\ref{sec:appendix_clean_only}).

When considering the 3 finetuned models used, RoBERTa achieved the highest accuracy across all datasets. Considering this and the much more similar performance of the established and newer LLM-based methods for RoBERTa as seen in Figure~\ref{fig:mean_diff_perf_para_vs_cont_ins}, this could be  indicating that even much cheaper established methods can achieve competitive model accuracy when compared to newer LLM-based augmentation methods on the best performing classifier, with only exceptions for a small number of seeds per label.

We also did a combination of the \emph{contextual insert} and \emph{backtranslation} methods as the two best established augmentation methods and compared it with the LLM-based methods, which did not result in a considerable increase in model precision compared to the \emph{paraphrasing} method. Details of this comparison can be found in the Appendix~\ref{sec:appendix_combination}.

We answer \emph{RQ1} as follows: in most cases, the established \emph{contextual word insert} augmentation has a better or similar effect on classifier accuracy than the LLM-based \emph{paraphrasing} augmentation. LLM mehtods perform better only with a small number of seeds per label. With increasing number of seeds per label, the difference between the two methods for accuracy starts to diminish.

\subsection{Analysis of Augmentation Costs and Benefits for Classifier Accuracy (RQ2)}\label{sec:analysis_cost_benefit}

\begin{table}[!t]
\centering
\small
\setlength\tabcolsep{3pt}
\begin{tabular}{@{}lcccccc@{}}
\toprule
 \emph{method} & \emph{Time cost} & \emph{kgCO$_2$ emitted} & \emph{Monetary cost} \\ \midrule
\emph{Backtrans.} & 46m 40s & 0.09 & \(\sim \)\$3  \\
\emph{Con. swap} & 36m 40s & 0.047 & \(\sim \)\$0.3   \\
\emph{Con. insert} & 40m & 0.047 & \(\sim \)\$0.3   \\
\emph{Para. LLM} & 1h 10m & 0.13 & \(\sim \)\$5  \\
\emph{Swap LLM} & 1h 10m & 0.13 & \(\sim \)\$5  \\
\emph{Insert LLM} & 1h 10m & 0.13 & \(\sim \)\$5  \\ \bottomrule
\end{tabular}
\caption{Approximated kgCO$_2$ emitted, time and monetary costs for each augmentation method on our hardware setup when collecting 15 samples for a 100 of seeds per label. The established methods take considerably less time and money while emitting far fewer emissions than newer LLM-based methods.}
\label{tab:approx_time_cost}
\end{table}

\begin{figure*}[!t]
\begin{tabular}{ccc}
  \includegraphics[width=0.3\textwidth]{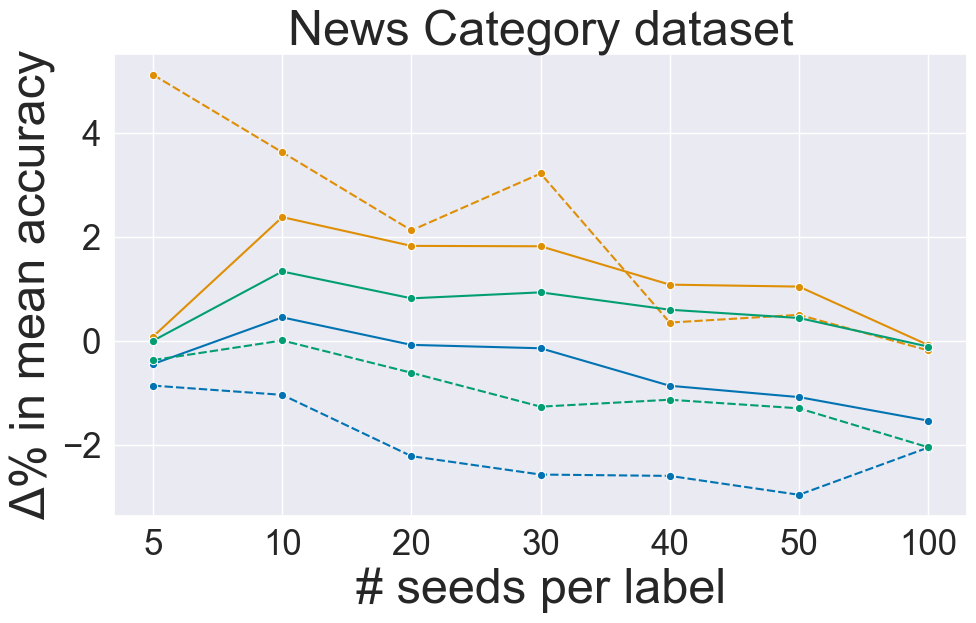} &   \includegraphics[width=0.3\textwidth]{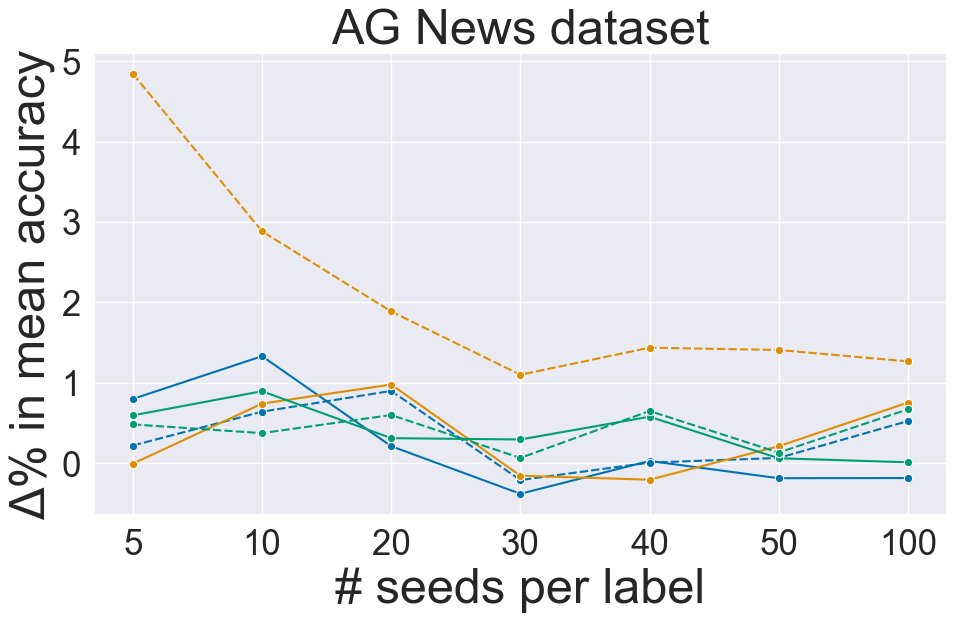}  & \includegraphics[width=0.3\textwidth]{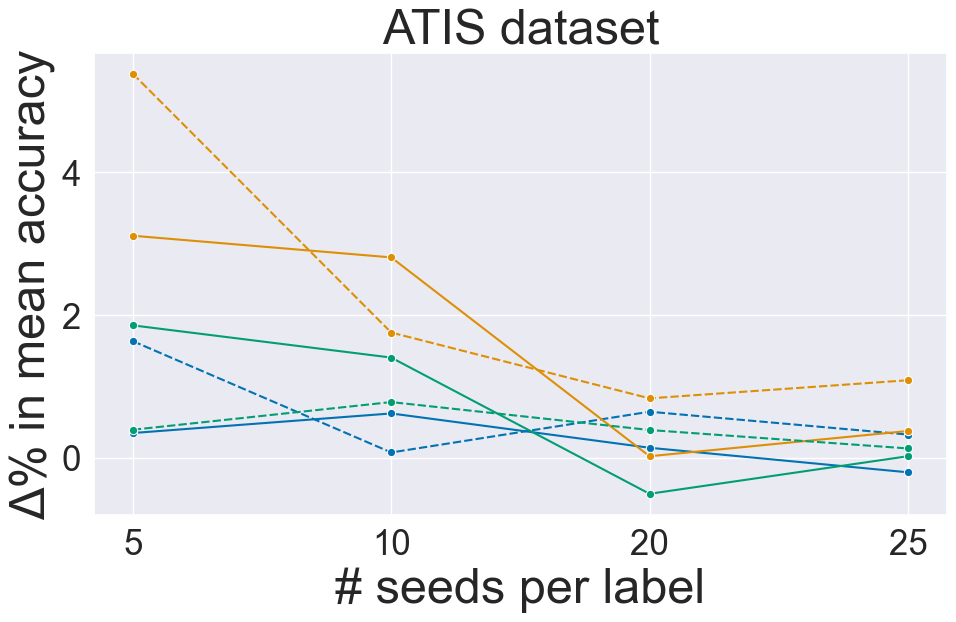} \\
  \includegraphics[width=0.3\textwidth]{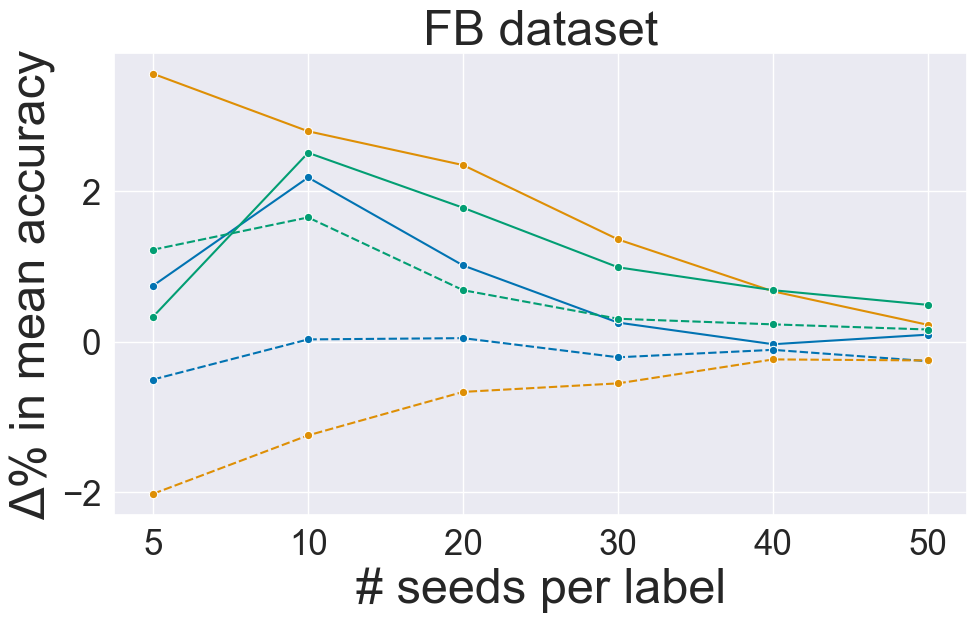} & \includegraphics[width=0.3\textwidth]{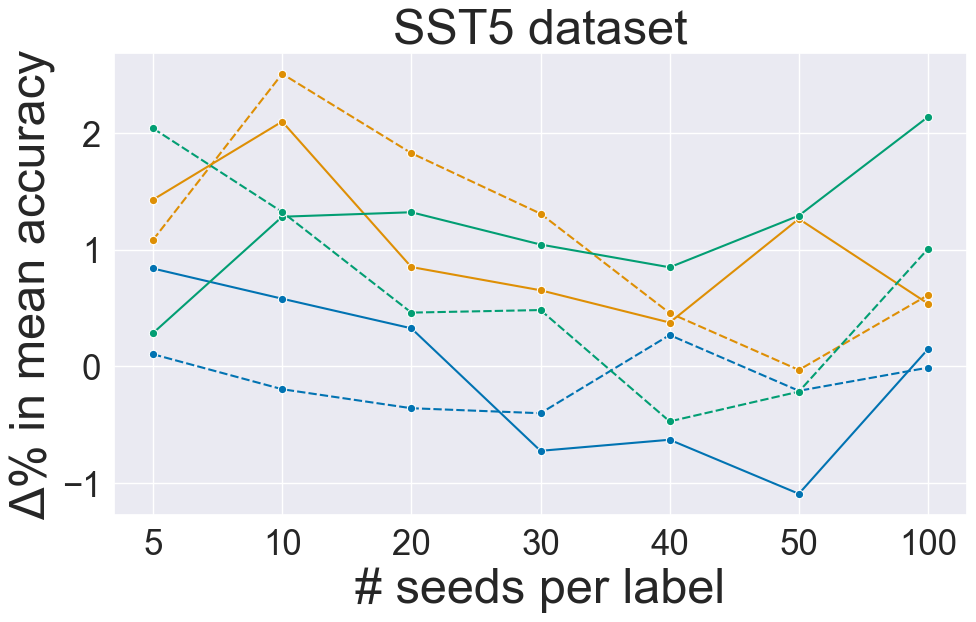} &   \includegraphics[width=0.3\textwidth]{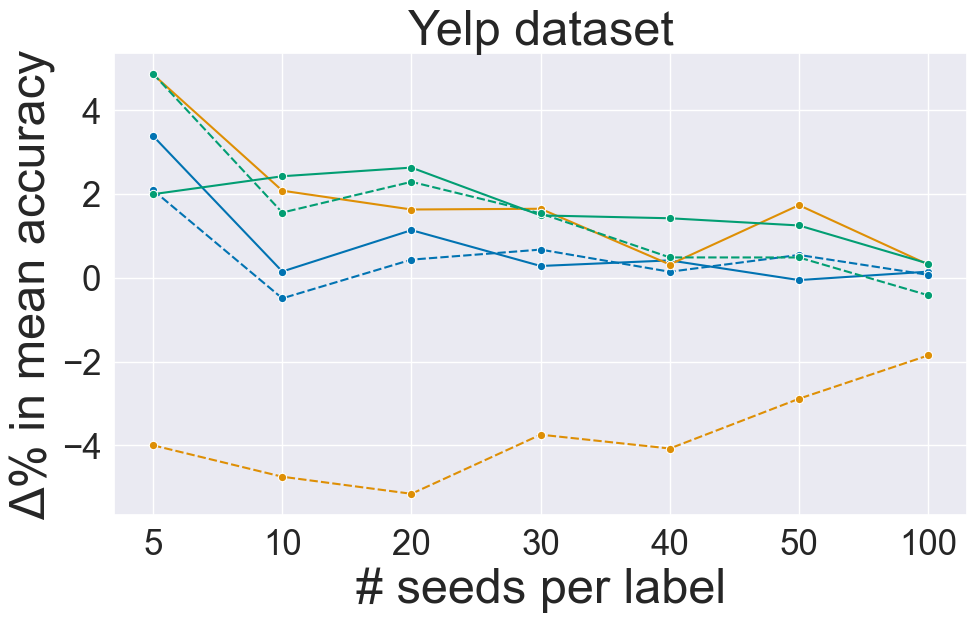} \\
  \multicolumn{3}{c}{\includegraphics[width=0.5\textwidth]{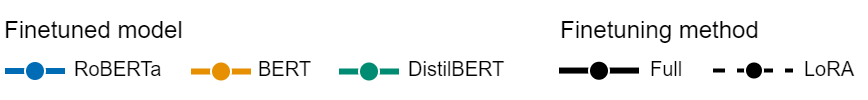}}
\end{tabular}
\caption{The difference in mean accuracy for classifiers trained on the \emph{paraphrasing} augmentation method and the \emph{contextual insert} augmentation method for 6 different datasets. The \emph{paraphrasing} method works generally better for a small (5-20) number of seeds per label and this benefit deteriorates with increased number of seeds per label.}
\label{fig:mean_diff_perf_para_vs_cont_ins}
\end{figure*}


To answer the RQ2, we first performed an approximate cost calculation for each of the used augmentation methods in terms of time needed to collect samples, monetary costs needed and emissions. We then identified cases where the higher cost of LLM-based methods is worth for the increased accuracy of classifiers.

We measure the time needed to collect the given number of samples on our hardware that we used for the experiments. We measured the time needed for collecting 15 augmented samples per 100 seed samples per label. The newer LLM-based methods have the same estimated time needed for data augmentation for each of the methods as we did not measure any significant differences in time needed between them. The results are displayed in Table~\ref{tab:approx_time_cost}.

Considering time only, the established augmentation methods run approximately 33\%-47\% faster than the newer LLM-based methods and when considering also the CO$_2$ emissions the \emph{contextual swap} and \emph{contextual insert methods}  emit approximately 64\% less kgCO$_2$ emissions for the same number of seed samples per label and number of collected samples. The details of how the  emissions approximation calculation was done can be found in Appendix~\ref{sec:appendix_emission_calc}. In terms of monetary costs, the \textit{context swap} and \textit{context insert} methods are approximately 16 times cheaper than the LLM-based methods. As such, the established text augmentation methods are considerably more efficient both in time needed per collected sample, monetary cost and in kgCO$_2$ emissions.

When considering the results in Figure~\ref{fig:mean_diff_perf_para_vs_cont_ins}, we observed increases of relative model classifier accuracy on small (5-20) number of seed samples when using the \emph{paraphrasing} method compared to the \emph{contextual insert} method is 3\%-17\%  better accuracy for classifiers when fine-tuning using QLoRA and 2\%-11\% when using full fine-tuning. However, for larger (30+) number of seed samples per label the positive relative increase range decreases for QLoRA  to 1.5\%-6\% and 0.5\%-4\% for full fine-tuning with a general increase of cases where the \emph{contextual insert} method performed better for classifier accuracy. Although the differences in the relative increase in model performance when using \textit{paraphrasing} method instead of \textit{context insert} method decrease for a higher number of seeds, the difference in costs and emissions increases. This is most evident for RoBERTa, which had the smallest relative increase in accuracy when using the \textit{paraphrasing} method instead of the \textit{context insert} method out of all of the finetuned models, with some benefits only for small number of seed samples per label used.

We answer \emph{RQ2} as follows: Considering the results of increased classifier accuracy trained on the \emph{paraphrasing} method augmentations for 5-20 per label against those trained on the \emph{contextual insert} method augmentations, the decreasing difference in accuracy between the methods with increasing number of seed samples per label and the augmentation methods cost approximation, the difference in accuracy seems to be worth the increased costs only for a small number of seeds. This is true for both full and QLoRA fine-tuning of models, while the difference in accuracy between the methods decreases significantly when using 30 seeds per label and more. Additionally, the cases where the \emph{contextual insert} method is better for model accuracy increase with more seeds per label used.

\begin{table*}[t!]
    \centering
    \small
    \setlength\tabcolsep{5pt}
    \begin{tabular}{@{}lc||c||c||c||c||c@{}}
    \toprule
    \textsc{Classifier}$\rightarrow$ & \multicolumn{2}{c}{\textsc{RoBERTa}} & \multicolumn{2}{c}{\textsc{BERT}} & \multicolumn{2}{c}{\textsc{DistilBERT}}\\
    Dataset$\downarrow$  & Full & QLoRA & Full & QLoRA & Full & QLoRA\\
     \midrule
    AG News & \texttt{~7 \textbf{\(|\)} 68 \textbf{\(|\)} ~1} & \texttt{13 \textbf{\(|\)} 50 \textbf{\(|\)} ~4} & \texttt{14 \textbf{\(|\)} 54 \textbf{\(|\)} ~1} & \texttt{24 \textbf{\(|\)} 36 \textbf{\(|\)} ~0} & \texttt{~7 \textbf{\(|\)} 70 \textbf{\(|\)} ~0} & \texttt{17 \textbf{\(|\)} 44 \textbf{\(|\)} ~3} \\
    News Category & \texttt{~2 \textbf{\(|\)} 58 \textbf{\(|\)} 11} & \texttt{~1 \textbf{\(|\)} 12 \textbf{\(|\)} 35} & \texttt{19 \textbf{\(|\)} 44 \textbf{\(|\)} ~1} & \texttt{19 \textbf{\(|\)} 38 \textbf{\(|\)} ~4} & \texttt{15 \textbf{\(|\)} 52 \textbf{\(|\)} ~1} & \texttt{~3 \textbf{\(|\)} 18 \textbf{\(|\)} 30} \\
    ATIS & \texttt{~4 \textbf{\(|\)} 36 \textbf{\(|\)} ~2} & \texttt{~7 \textbf{\(|\)} 22 \textbf{\(|\)} ~6} & \texttt{~9 \textbf{\(|\)} 24 \textbf{\(|\)} ~3} & \texttt{~9 \textbf{\(|\)} 30 \textbf{\(|\)} ~0} & \texttt{~9 \textbf{\(|\)} 16 \textbf{\(|\)} ~7} & \texttt{10 \textbf{\(|\)} 20 \textbf{\(|\)} ~4} \\
    FB & \texttt{20 \textbf{\(|\)} 28 \textbf{\(|\)} ~2} & \texttt{~5 \textbf{\(|\)} 42 \textbf{\(|\)} 10} & \texttt{25 \textbf{\(|\)} 22 \textbf{\(|\)} ~0} & \texttt{~3 \textbf{\(|\)} 28 \textbf{\(|\)} 19} & \texttt{25 \textbf{\(|\)} 20 \textbf{\(|\)} ~1} & \texttt{20 \textbf{\(|\)} 16 \textbf{\(|\)} ~8} \\
    SST-5 & \texttt{~7 \textbf{\(|\)} 58 \textbf{\(|\)} ~6} & \texttt{~8 \textbf{\(|\)} 46 \textbf{\(|\)} 11} & \texttt{15 \textbf{\(|\)} 52 \textbf{\(|\)} ~1} & \texttt{15 \textbf{\(|\)} 50 \textbf{\(|\)} ~2} & \texttt{17 \textbf{\(|\)} 50 \textbf{\(|\)} ~0} & \texttt{19 \textbf{\(|\)} 36 \textbf{\(|\)} ~5} \\
    Yelp & \texttt{11 \textbf{\(|\)} 56 \textbf{\(|\)} ~3} & \texttt{15 \textbf{\(|\)} 42 \textbf{\(|\)} ~6} & \texttt{19 \textbf{\(|\)} 42 \textbf{\(|\)} ~2} & \texttt{~0 \textbf{\(|\)} ~8 \textbf{\(|\)} 38} & \texttt{17 \textbf{\(|\)} 48 \textbf{\(|\)} ~1} & \texttt{31 \textbf{\(|\)} 12 \textbf{\(|\)} ~5} \\ \bottomrule
    \end{tabular}
    \caption{Comparison of the number of cases where models trained using data from either \emph{paraphrasing} or \emph{contextual insert} methods worked statistically \textit{(p=0.05)} better or had similar accuracy when compared between each other. The numbers represent the result of one statistical test between 10 fine-tunings of the given model on data collected via the \emph{paraphrasing} or the \emph{contextual insert} using a specific random seed for a given number of seed samples per label. The cells are formatted in this way: \texttt{[\char"0023~paraphrasing was better]} \textbf{\(|\)} \texttt{[\char"0023~similar accuracy]} \textbf{\(|\)} \texttt{[\char"0023~contextual insert was better]}. In most cases, the \emph{paraphrasing} method works better for BERT and DistilBERT in both full fine-tuning and QLoRA fine-tuning.}
    \label{tab:count_perf_para_vs_cont_ins}
\end{table*}

\section{Discussion}
 The results of our experiments lead to the following observations: First, the \emph{paraphrasing} method was the best within the newer LLM-based augmentation methods, considering classifier accuracy. This could be due to the demonstrated ability~\cite{cegin-etal-2023-chatgpt} of the newer LLMs to create very diverse paraphrases, being less constrained by seed samples. The \emph{contextual insert} method worked best within the established augmentation methodscaused by the \emph{backtranslation} method creating a lot of duplicated samples, and the \emph{contextual swap} method introducing less variety than the \emph{contextual insert} method.
 
 Second, the number of cases in which the \emph{paraphrasing} method as (a LLM-based method) significantly outperforms the established \emph{contextual insert} methodth more seed samples per label. This is similar to previous studies~\cite{dai2023auggpt, ubani2023zeroshotdataaug}, as we observed this in nearly all the cases for a small number of seeds. This finding differs from the results of the previous study~\cite{piedboeuf-langlais-2023-chatgpt} , where such increase was observed less often. A different number of collected samples, classifiers used and other factors might be the reason for this disparity. \textbf{Generally, the LLM-based methods achieve better classifier accuracy than established methods in the cases of very small seed numbers, which points to their potential benefits in low-resource settings}.

 Third, furthermore, when we increased the number of seeds, we observed a decrease of accuracy differences between models trained on data from \emph{paraphrasing} and models trained with \emph{contextual insert}, similar to~\cite{piedboeuf-langlais-2023-chatgpt}. The highest relative increase in model accuracy with\emph{paraphrasing}  instead of \emph{contextual insert} appears with 5 to 20 seeds per label. After 30 or more seeds were used,  the relative difference between methods decreased. Additionally, the difference between the LLM-based and established methods in terms of monetary costs, time costs, and emissions is quite significant (see section~\ref{sec:analysis_cost_benefit}). \textbf{Therefore, it seems beneficial,  from the perspective of both cost and model accuracy to use the newer LLM-based augmentation methods only in low-resource settings}.

 Fourth, we observed some exceptions to the trends reviewed above. \textbf{The fine-tuned RoBERTa models (which provided best classification accuracy among the fine-tuned models) generally benefited more from augmentation methods which used insertion of words}. This might be due to a more robust pretraining of RoBERTa, where augmentations that introduce more noise, are less beneficial for training. Another case was the fine-tuning of BERT using QLoRA, where for some cases the \emph{paraphrasing} method was either considerably better or worse than the \emph{contextual insert} method for classifier accuracy. This might be due to differences in the pre-training data and processes used for BERT in comparison with DistilBERT or RoBERTa, making it far more sensible to text augmentation methods when using QLoRA.

 Fifth, the difference between the \emph{paraphrasing} and the \emph{contextual insert} method on model accuracy had much more variance for QLoRA than for full fine-tuning. When the \emph{paraphrasing} method is used for QLoRA on classifiers, the increased accuracy (compared to \emph{contextual insert}) is generally smaller than with full fine-tuning. \textbf{LLM paraphrasing's sample variability might be providing more benefits when the model can leverage it through full fine-tuning.} For QLoRA finetuning, both methods provide enough variability to improve accuracy without significant differences.

 Sixth, \textbf{the combination of the best established methods does not improve their overall accuracy of the downstream model compared to only using \emph{contextual insert}}. This might be due to the combination of methods leading to a possible distribution shift or models overfitting on the augmented data.

To summarize, s the costs of using established augmentation methods is considerably lower than the newer LLM-based methods, it appears to be beneficial to use them instead of newer LLM methods for higher number of seeds per label when targeting model accuracy and use LLM-methods in cases of low-resource setting where the relative gain in accuracy is highest.
 
\section{Conclusion}

We compared the effects of newer \emph{LLM-based} and \emph{established} textual augmentation methods on downstream classifier accuracy. We compared them for combinations of 6 datasets, 3 classifiers, 2 fine-tuning approaches, 2 augmenting LLMs, various numbers of seed samples per label and numbers of augmented samples per seed. In total, we analysed a total of 267,300 fine-tunings. We identified the \emph{paraphrasing} method as the best performing LLM-based and the \emph{contextual insert} as the best performing established augmentation method. The comparison of these two best methods indicates that the use of LLM-based methods for data augmentation, instead of established methods, is only warranted for a small number of seed samples per label (5 to 20). There, we observed a statistically significant increase of cases where LLM-based methods are better and observed higher relative increases of model accuracy compared to established methods. However, with increasing number of seeds per label, this effect decreased, and the number of cases of established methods having a higher influence on the accuracy of classifiers increased. As newer LLM methods are considerably more costly than established methods, their use is justified only for low-resource settings, where differences between the method's costs are smaller.


\section*{Limitations}
    We note several limitations to our work. 
    
    First, we only used datasets, augmentation methods, and LLMs for the English language and did not investigate cases of multi-lingual text augmentation.

    Second, we did not use various patterns of prompts  and followed those used in previous studies~\cite{cegin-etal-2023-chatgpt, larson-etal-2020-iterative}. Different prompts could have effects on the quality of text augmentataions, but they would also radically increase the size of this study, and thus, we decided to leave this for future work.

    Third, we did not use newer LLMs for classification fine-tuning via PEFT methods (e.g fine-tuning of Llama-3 or Mistral using QLoRA). While such inclusion would strengthen our findings, we decided not to use these models for classification fine-tuning due two main reasons. First, the evaluation of these models is very costly and takes a long time due to their size, which results in them being mostly used with a small subset of the testing data~\cite{chang-jia-2023-data, li-qiu-2023-finding, gao-etal-2021-making, koksal-etal-2023-meal}. This, in return, can lead to unintentionally cherry picked results. Second, to do an analysis of this size for the combinations of parameters that influence one fine-tuning of models, we had to do a total of 44,500 fine-tuning for one model and fine-tuning method combination. A fine-tuning 44,500 times of a smaller generative LLM with 7B parameters and then evaluating it on a substantial split of the test data was infeasible to us time- and cost-wise. It would also radically increase the energy consumption of this study, and in turn emissions emitted. 

    Fourth, from the family of PEFT methods we used only QLoRA and not multiple different PEFT methods. We opted for QLoRA due to its popularity and good performance. While including more fine-tuning methods in the paper would increase the strength of the findings and provide an even finer analysis of cases, it would also, similar to the case of not fine-tuning LLMs for classification from the previous limitation, lead to a significantly higher number of fine-tunings needed for a proper analysis of the new fine-tuning method added.

    Fifth, for the LLM augmentation methods we used only Llama-3-8B and GPT-3.5. We did not use larger models (e.g. 70B or GPT-4) as their increased performance in text augmentation for model accuracy has been shown~\cite{cegin2024effectsdiversityincentivessample} to be not that significant when compared to variants of LLMs with fewer parameters.  

    Sixth, we only used 3 established methods compared to previous studies~\cite{piedboeuf-langlais-2023-chatgpt, dai2023auggpt, ubani2023zeroshotdataaug}, which used more established methods for their comparisons. In our case, we used different types of methods which had a good performance in a previous study~\cite{dai2023auggpt}. While inclusion of multiple other established methods would increase the strength of our findings, it would also require a lot additional fine-tunings and evaluation to be done in order to get results for our detailed analysis.

    Seventh, we did not enhance the LLM-based methods of \textit{word insertion} or \textit{word swap} with heuristics to select locations in seed texts where words should be replaced or added. We opted against this to let the LLMs decide internally (as a blackbox) which words to replace or add and where, providing these methods with simplicity and without additional potential costs. A potential extension of these LLM-methods with heuristics of where to replace or add words could possibly improve the performance of these methods for augmentation, and we see this as a natural extension of our work.

    Eight, the \textit{backtranslation} method could be improved by adding multiple languages into the translation process which would possibly increase the lexical diversity of and number of created paraphrases. However, this would also increase the cost of using this method, which is already the most costly from all of the established augmentation methods.

    Ninth, we only focus on classification tasks and make no claims about the effects of established and LLM-based text augmentation on other NLP tasks.

    Tenth, we do not know if any of the 6 datasets used in this study have been used for training the LLMs we used for data collection and if this had any effect on our results and findings. As such, we do not know how much would be the comparison of established and newer LLM augmentation methods different on new, unpublished datasets. This limitation is part of the recently recognized possible ``LLM validation crisis'', as described by~\cite{li2023task}.

\section*{Acknowledgments}

This work was partially supported by AI-CODE - AI services for COntinuous trust in emerging Digital Environments, a project funded by the European Union under the Horizon Europe, GA No. 101135437. https://doi.org/10.3030/101135437 ; and Modermed, a project funded by the Slovak Research and Development Agency, GA No. APVV-22-0414. The work was also partially funded by the EU NextGenerationEU through the Recovery and Resilience Plan for Slovakia under the projects No. 09I03-03-V03-00020 and No. 09I01-03-V04-00006.

Part of the research results were obtained using the computational resources procured in the national project National competence centre for high performance computing (project code: 311070AKF2) funded by European Regional Development Fund, EU Structural Funds Informatization of society, Operational Program Integrated Infrastructure. This work was supported by the Ministry of Education, Youth and Sports of the Czech Republic through the e-INFRA CZ (ID:90254).
\bibliography{compressed}

\appendix

\section{Ethical considerations}
Based on a thorough ethical assessment, performed on the basis of intra-institutional ethical guidelines and checklists tailored to the use of data and algorithms, we see no ethical concerns pertaining directly to the conduct of this research. We also ethically assessed our paraphrase validity crowdsourcing process from Appendix~\ref{sec:appendix_para_valid} via our intra-institutional ethical guidelines and found no ethical concerns. In our study, we analysed existing data or data generated using various LLMs. During our manual checking of the data we also ensured that the data contained no personal or offensive data. Albeit production of new data through LLMs bears several risks, such as introduction of biases, the small size of the produced dataset, sufficient for experimentation is, at the same time, insufficient for any major machine learning endeavors, where such biases could be transferred.

We follow the license terms for all the models and datasets we used (such as the one required for the use of the Llama-3 model) – all models and datasets allow their use as part of research.

\section{Details of CO2 Emission calculation and emissions related to experiments}\label{sec:appendix_emission_calc}
For the estimations we used the \href{https://mlco2.github.io/impact#compute}{MachineLearning Impact calculator} presented in \cite{lacoste2019quantifying}.
For estimations of GPU emissions we used hardware of type A100 PCIe 40/80GB (TDP of 250W) and for estimation of CPU emissions we used hardware of type Intel Xeon Gold 6148. 

We conducted the data collection and fine-tuning on a custom private infrastructure with 16 core CPU, 64 GB RAM and 4xA100 GPUs. For the LLM-based augmentation methods and \textit{backtranslation} method we used the GPU to collect data, while for the \textit{context insert} and \textit{context swap} methods we used CPUs only.

Data collection via Llama-3 was conducted using a private infrastructure, which has a carbon efficiency of 0.432 kg CO$_2$/kWh. A cumulative of 20 GPU hours of computation was performed on hardware of type A100 PCIe 40/80GB (TDP of 250W) for data collection.

Model fine-tuning for both all of the fine-tuned models using either full fine-tuning or QLoRA fine-tuning was conducted using a private infrastructure, which has a carbon efficiency of 0.432 kg CO$_2$/kWh. Approximately a cumulative of 1100 GPU hours of computation was performed on hardware of type A100 PCIe 40/80GB (TDP of 250W) for data collection.

Total emissions together are estimated to be 120.96 kgCO$_2$ of which 0 percents were directly offset. We tried to reduce the generated emissions by using 4-bit quantization for Llama-3 data collection and QLoRA training.

\section{Augmented samples validity: checking process and results}\label{sec:appendix_para_valid}

For the process of checking the validity of the created augmented samples, we used our very own web app developed for this process. The users, who were the authors that also developed the app, were shown the seed samples, its label and one particular sample to validate. The authors/users all gave consent to the data collection process and had knowledge of how the data would be used. The instructions were \emph{"Please decide if the augmented sample has the same meaning as the seed sentence and if it adheres to the label of the seed sentence."}  The user was then able to either mark the sample as valid or not, with an additional optional checkbox to label the samples as ‘borderline case’ for possible re-visions. As the seed sentence changed only once in a while (we first showed all the paraphrases from one seed sentence) this significantly reduced the cognitive load on the annotator. The users/authors then discussed together the ‘borderline cases’ where the users were not sure about the validity of created paraphrases.

Before evaluating validity of each augmentation method and the samples it produces, we filtered for malformed augmented samples, empty samples or duplicated samples as per~\cite{cegin-etal-2023-chatgpt}. There were no such samples detected for the newer LLM-based methods. We detected around 0.05\%-0.5\% of all augmented samples to be duplicates for the \emph{contextual word insertion} and \emph{contextual word replacement}. The worst number of duplicates was detected for the \emph{backtranslation} method with 80\% of all collected augmented samples to be duplicates. This still meant that we collected at least 2 to 3 augmented samples per seed and as such we did not eliminate this method from further evaluation. For fine-tuning cases using the \emph{backtranslation} method where more number of collected samples per seed than 3 were needed we used all of the available collected unique augmented samples. This high number of duplicates might be due to the translation model limitations with repeating patterns, as well as using only one intermediary language as per the original paper.

\section{Effects of number of collected augmented samples from augmentation methods on model accuracy}
\label{sec:appendix_effects_of_no_col_samples}

\begin{figure*}[!t]
\begin{tabular}{cc}
  \includegraphics[width=0.475\textwidth]{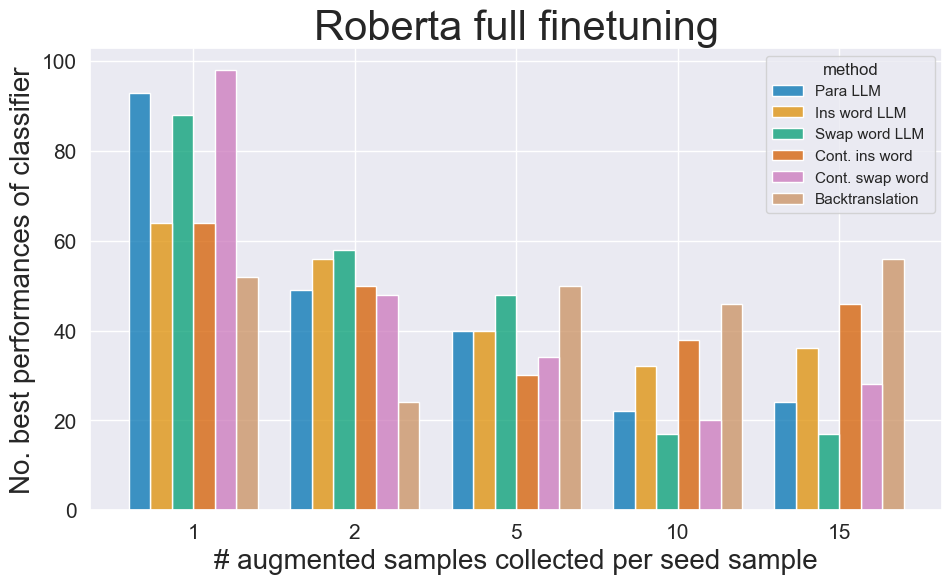} &   \includegraphics[width=0.475\textwidth]{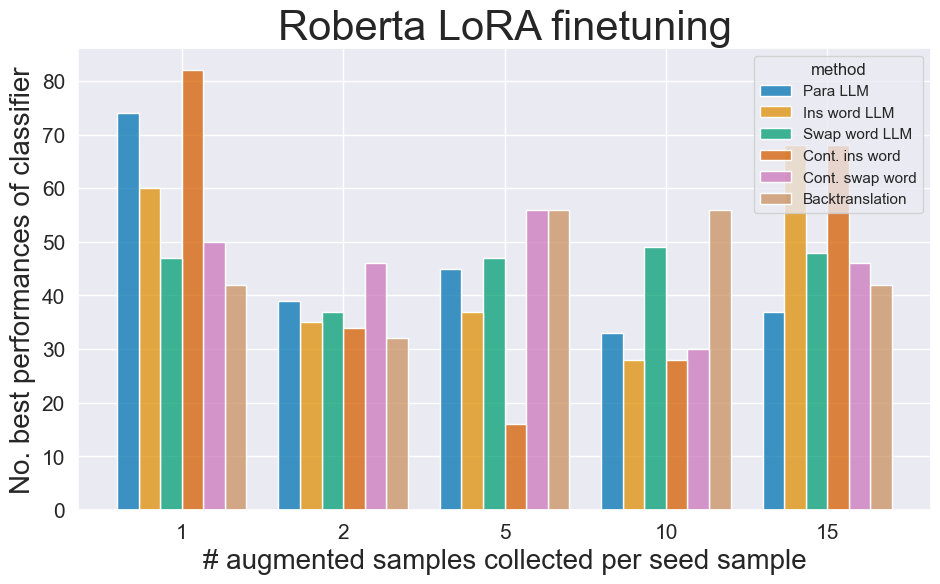} \\
 \includegraphics[width=0.475\textwidth]{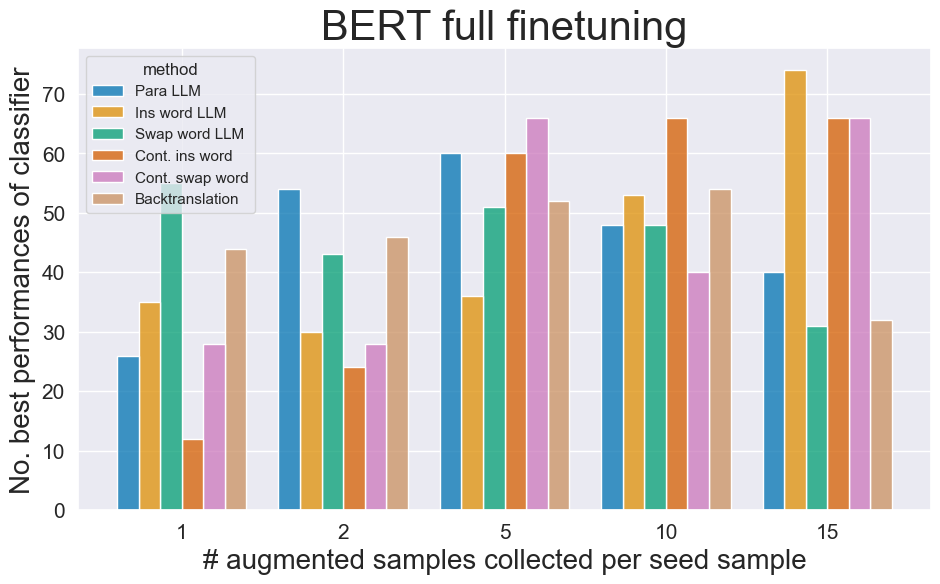} &   \includegraphics[width=0.475\textwidth]{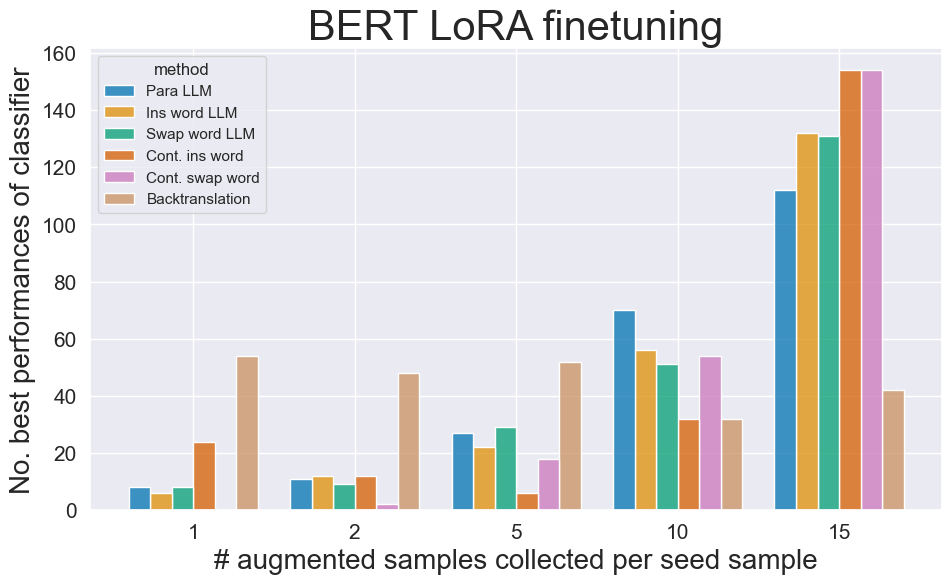} \\
 \includegraphics[width=0.475\textwidth]{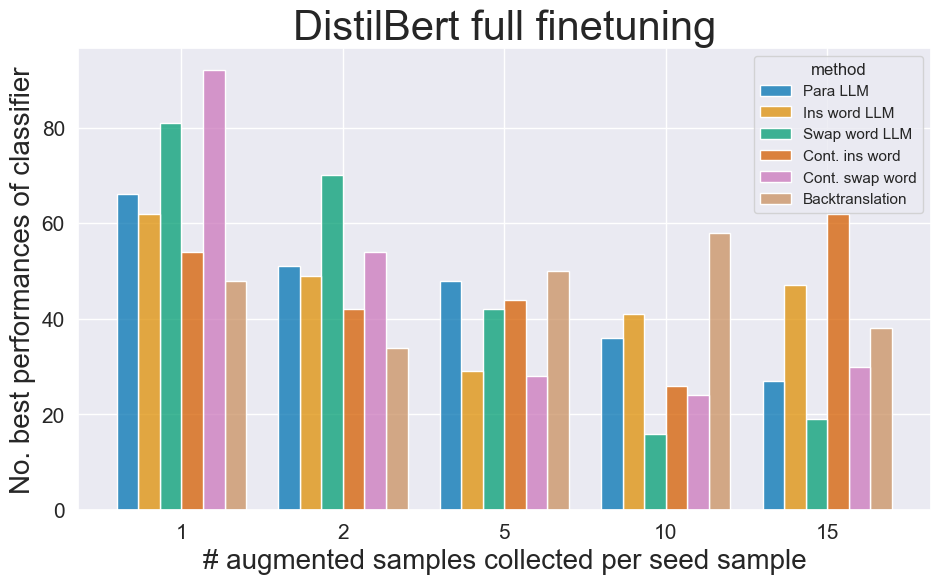} &   \includegraphics[width=0.475\textwidth]{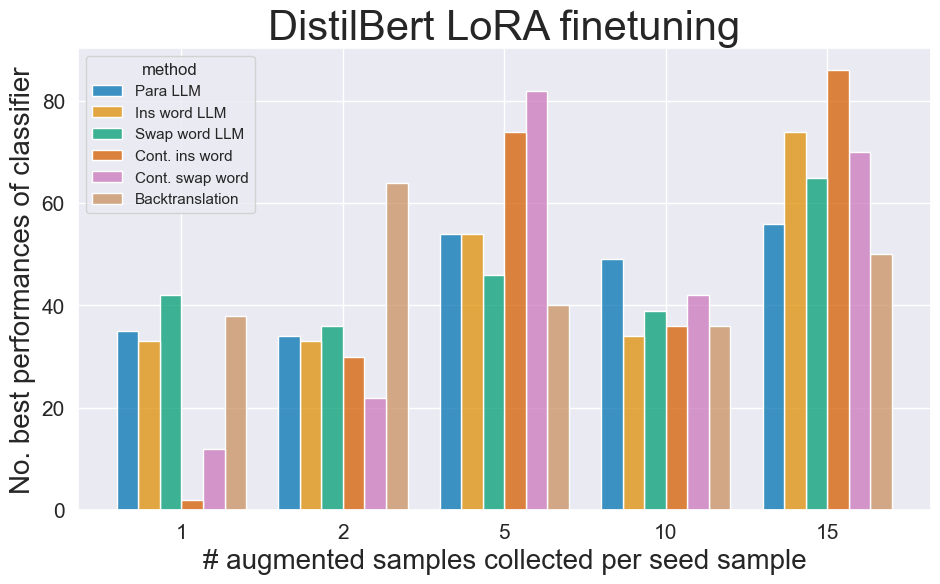} \\
\end{tabular}
\caption{The number of cases per number of collected augmented samples per seed sample where each augmentation method achieved best accuracy for 6 different combinations of models and fine-tuning methods. Except for RoBERTa and DistilBERT full fine-tuning, the methods worked best for model accuracy when more augmented samples were provided.}
\label{fig:effect_no_coll_on_perf}
\end{figure*}

 In this section we compare the effects of number of collected augmented samples per seed sample have on model accuracy. We noticed that QLoRA fine-tuning benefited from more collected augmented samples per seed sample than full fine-tuning of classifiers for all of the methods. RoBERTa and DistilBERT full fine-tuning generally needed only few (less than 5) augmented samples per seed sample to achieve best classifier accuracy across different augmentation methods. This might be due to the more robust pretraining process in case of RoBERTa and the distillation training of DistilBERT, where pretrained weights of the models benefit less from more added noise to the dataset via increased number of collected augmented samples. Bert full fine-tuning had a similar trend as RoBERTa and DistilBERT with the exception of inserting words based methods. This might indicate that while a lot of noise might degrade the accuracy of fine-tuned Bert (as is the case for the \emph{paraphrasing} method), the augmented samples from word insertion add just enough noise for the model to benefit from it. Additionally with increased number of seed samples we observed that less number of augmented samples per seed sample were needed for the fine-tuned models to achieve best accuracy, indicating that a lot of augmented examples in the training data could lead to a distribution shift. Visualization of these results can be found in Figure~\ref{fig:effect_no_coll_on_perf}.

\section{Combination of best established augmentation methods for classifier accuracy}\label{sec:appendix_combination}

Given that the established augmentation methods are cheaper when compared to newer LLM-based augmentation methods, we can combine the established augmentation methods together and then compare them with newer LLM-based methods, specifically the \emph{paraphrasing method}, to determine if such combination increases accuracy of models for classification.

To do so, we combine the \emph{backtranslation} and \emph{contextual insert} method in this way: for each number of collected samples per seed sample we use from both methods augmented samples, e.g. for 2 number of collected samples used per seed sample from the \emph{paraphrasing} method we include 2 from the \emph{backtranslation} method and 2 from the \emph{contextual insert} method. As mentioned in Appendix~\ref{sec:appendix_para_valid} the \emph{backtranslation} method produces a lot of duplicates samples, meaning that for cases where not enough unique augmented samples are collected we used all of the available augmented samples.

The comparison of the combination of the established methods and the \emph{paraphrasing} method for classifier accuracy can be seen in Table~\ref{tab:count_perf_para_vs_comb}. Compared to the results from Section~\ref{sec:established_newer_strict_comp} where we compared the \emph{paraphrasing} method against only the \emph{contextual insert} method, the combination of the two best established methods yields more cases where the \emph{paraphrasing} method is better for all the fine-tuning methods and dataset combinations. This might be due to such combination of two augmentation methods introducing a potential distribution shift in the data, overfitting on augmented data or possible inconsistencies in the augmented data. To conclude, the combination of the established methods increases the cost of augmentation while providing worse results compared to only using the \emph{contextual insert} method for classifier accuracy.

\begin{table*}[t!]
    \centering
    \small
    \setlength\tabcolsep{5pt}
\begin{tabular}{@{}lc||c||c||c||c||c@{}}
\toprule
\textsc{Classifier}$\rightarrow$ & \multicolumn{2}{c}{\textsc{RoBERTa}} & \multicolumn{2}{c}{\textsc{BERT}} & \multicolumn{2}{c}{\textsc{DistilBERT}}\\
Dataset$\downarrow$  & Full & QLoRA & Full & QLoRA & Full & QLoRA\\
 \midrule
AG News & \texttt{~7 \textbf{\(|\)} 70 \textbf{\(|\)} ~0} & \texttt{16 \textbf{\(|\)} 46 \textbf{\(|\)} ~3} & \texttt{12 \textbf{\(|\)} 54 \textbf{\(|\)} ~3} & \texttt{30 \textbf{\(|\)} 24 \textbf{\(|\)} ~0} & \texttt{~8 \textbf{\(|\)} 62 \textbf{\(|\)} ~3} & \texttt{25 \textbf{\(|\)} 30 \textbf{\(|\)} ~2} \\
News Category & \texttt{~3 \textbf{\(|\)} 60 \textbf{\(|\)} ~9} & \texttt{~1 \textbf{\(|\)} 12 \textbf{\(|\)} 35} & \texttt{19 \textbf{\(|\)} 44 \textbf{\(|\)} ~1} & \texttt{30 \textbf{\(|\)} 24 \textbf{\(|\)} ~0} & \texttt{10 \textbf{\(|\)} 62 \textbf{\(|\)} ~1} & \texttt{~3 \textbf{\(|\)} 26 \textbf{\(|\)} 26} \\
ATIS & \texttt{12 \textbf{\(|\)} 18 \textbf{\(|\)} ~3} & \texttt{14 \textbf{\(|\)} 12 \textbf{\(|\)} ~4} & \texttt{10 \textbf{\(|\)} 26 \textbf{\(|\)} ~1} & \texttt{12 \textbf{\(|\)} 22 \textbf{\(|\)} ~1} & \texttt{12 \textbf{\(|\)} 16 \textbf{\(|\)} ~4} & \texttt{12 \textbf{\(|\)} 22 \textbf{\(|\)} ~1} \\
FB & \texttt{11 \textbf{\(|\)} 42 \textbf{\(|\)} ~4} & \texttt{11 \textbf{\(|\)} 32 \textbf{\(|\)} ~9} & \texttt{21 \textbf{\(|\)} 22 \textbf{\(|\)} ~4} & \texttt{36 \textbf{\(|\)} ~0 \textbf{\(|\)} ~0} & \texttt{16 \textbf{\(|\)} 32 \textbf{\(|\)} ~4} & \texttt{19 \textbf{\(|\)} 14 \textbf{\(|\)} 10} \\
SST-5 & \texttt{~5 \textbf{\(|\)} 72 \textbf{\(|\)} ~1} & \texttt{~5 \textbf{\(|\)} 48 \textbf{\(|\)} 13} & \texttt{~9 \textbf{\(|\)} 64 \textbf{\(|\)} ~1} & \texttt{21 \textbf{\(|\)} 42 \textbf{\(|\)} ~0} & \texttt{14 \textbf{\(|\)} 54 \textbf{\(|\)} ~1} & \texttt{20 \textbf{\(|\)} 26 \textbf{\(|\)} ~9} \\
YELP & \texttt{~7 \textbf{\(|\)} 58 \textbf{\(|\)} ~6} & \texttt{~9 \textbf{\(|\)} 46 \textbf{\(|\)} 10} & \texttt{16 \textbf{\(|\)} 52 \textbf{\(|\)} ~0} & \texttt{40 \textbf{\(|\)} ~4 \textbf{\(|\)} ~0} & \texttt{11 \textbf{\(|\)} 58 \textbf{\(|\)} ~2} & \texttt{26 \textbf{\(|\)} 20 \textbf{\(|\)} ~6} \\
\bottomrule
    \end{tabular}
    \caption{Comparison of the number of cases where models trained using data from either \emph{paraphrasing} or a combination of \emph{contextual insert} and \emph{backtranslation} methods worked statistically (p=0.05) better or had similar accuracy when compared between each other. The numbers represent the result of one statistical test between 10 fine-tunings of the given model on data collected via the \emph{paraphrasing} or the combination of two methods using a specific random seed for a given number of seed samples per label. The cells are formatted in this way: \texttt{[\char"0023~paraphrasing was better]} \textbf{\(|\)} \texttt{[\char"0023~similar accuracy]} \textbf{\(|\)} \texttt{[\char"0023~combination was better]}. In most cases the \emph{paraphrasing} method works better for BERT and DistilBERT in both full fine-tuning and QLoRA fine-tuning, with an decrease of such cases when fine-tuning RoBERTa.}
    \label{tab:count_perf_para_vs_comb}
\end{table*}

\section{Results for other combinations of LLM-based and established methods on model accuracy}\label{sec:appendix_other_model_perf_details}

\begin{figure*}[!t]
\begin{tabular}{ccc}
  \includegraphics[width=0.3\textwidth]{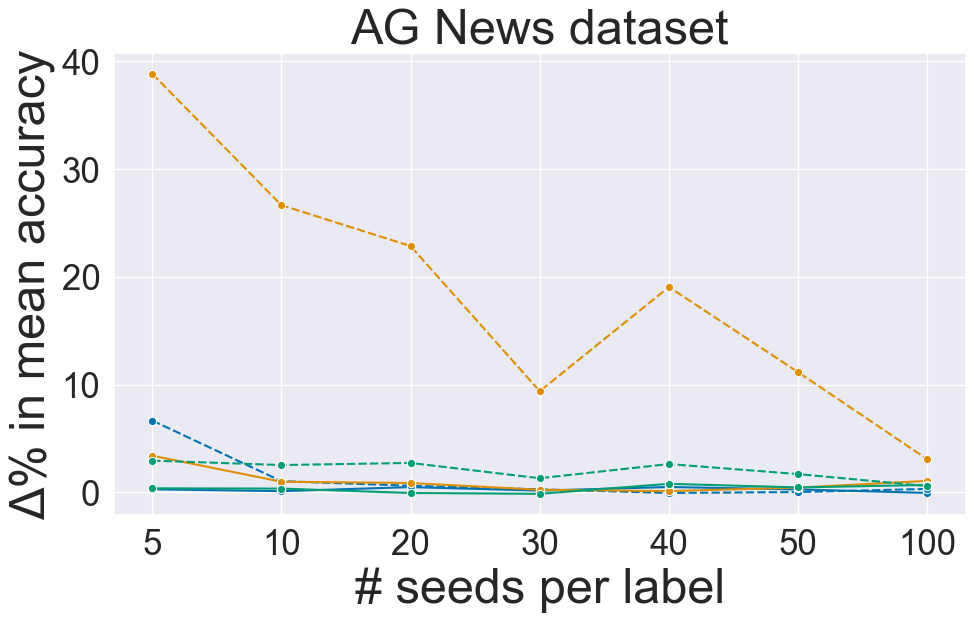} &   \includegraphics[width=0.3\textwidth]{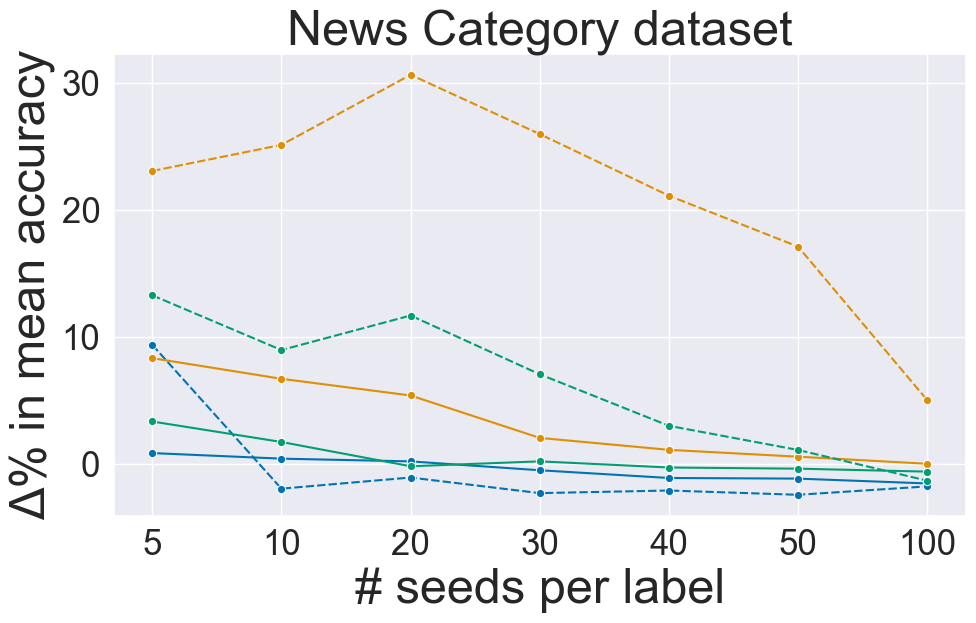}  & \includegraphics[width=0.3\textwidth]{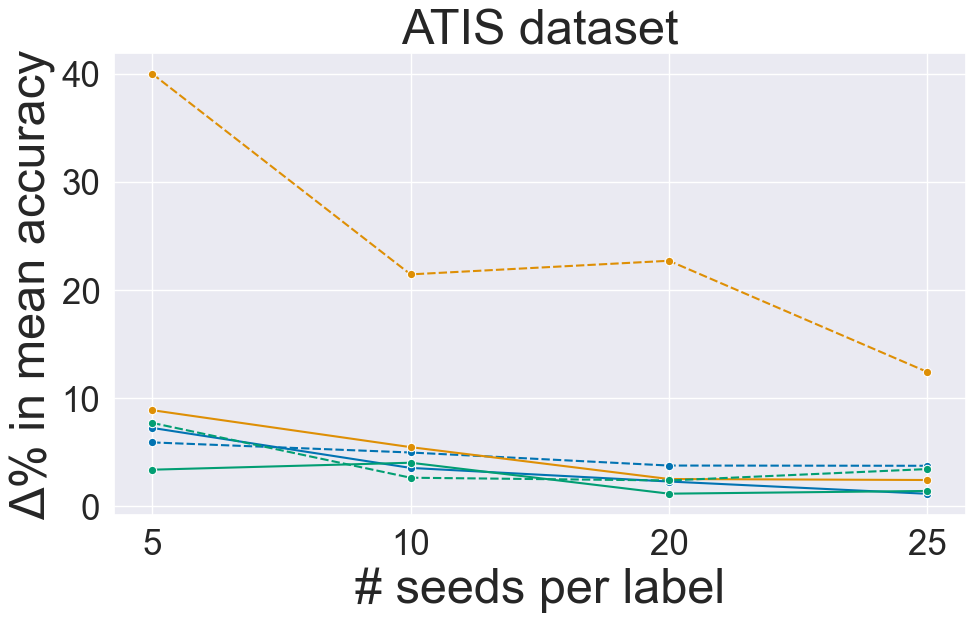} \\
  \includegraphics[width=0.3\textwidth]{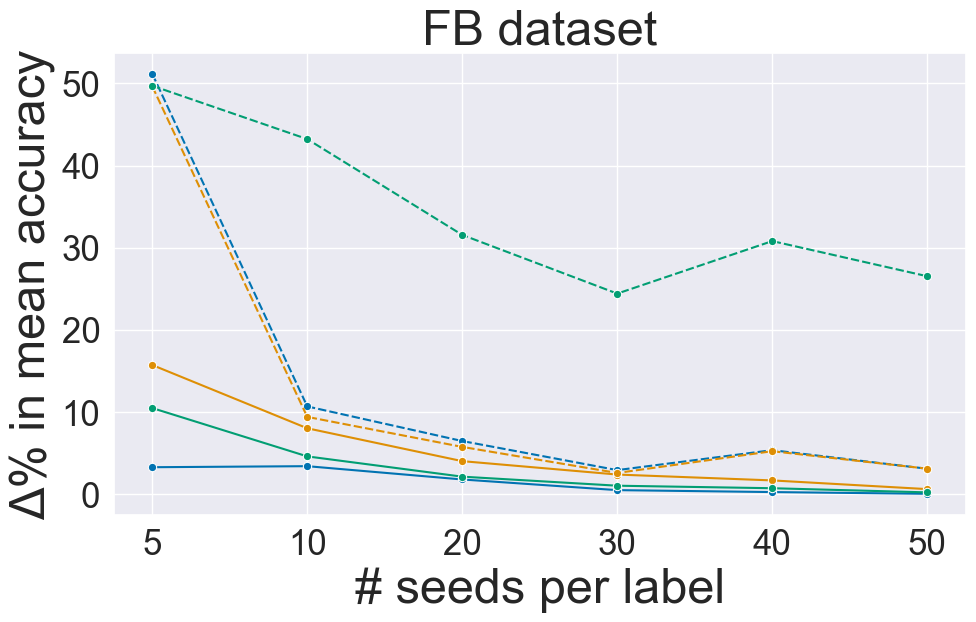} & \includegraphics[width=0.3\textwidth]{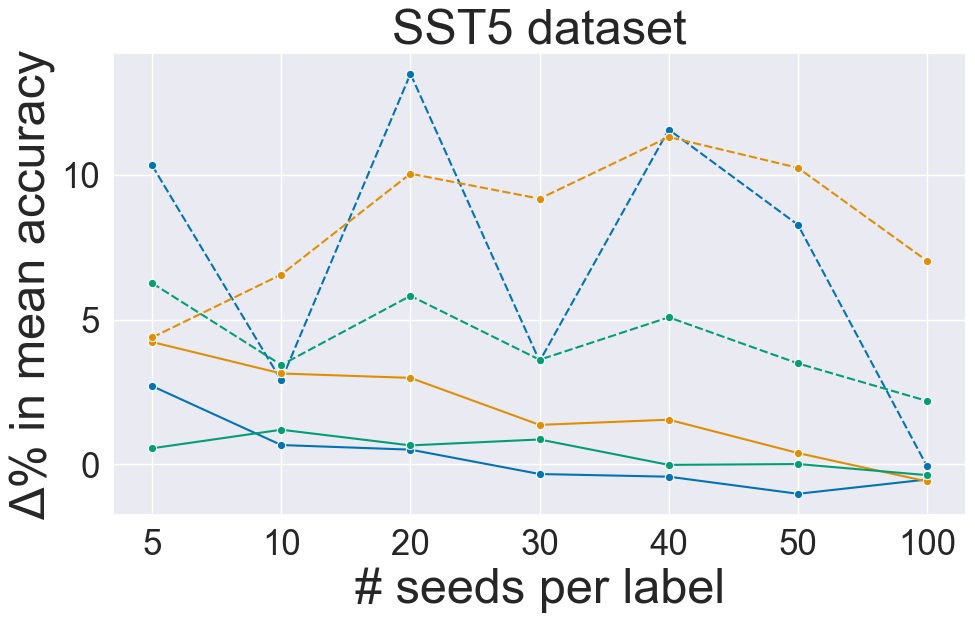} &   \includegraphics[width=0.3\textwidth]{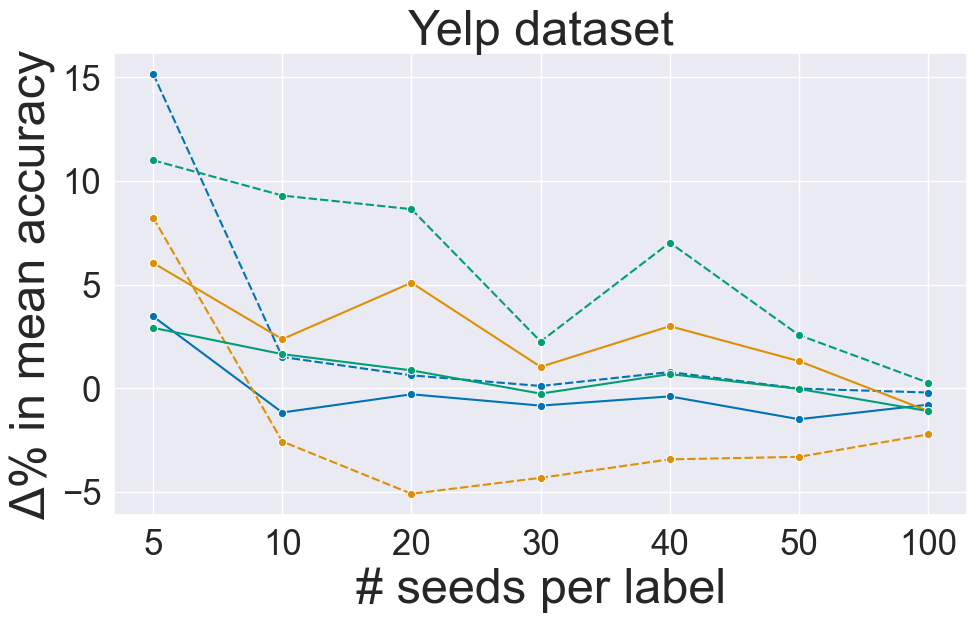} \\
  \multicolumn{3}{c}{\includegraphics[width=0.5\textwidth]{figures/legenda.png}}
\end{tabular}
\caption{The difference in mean accuracy for classifiers trained on the \emph{paraphrasing} augmentation method and the \emph{backtranslation} augmentation method for 6 different datasets. The \emph{paraphrasing} method works generally better in all cases.}
\label{fig:mean_diff_perf_para_vs_back}
\end{figure*}

\begin{figure*}[!t]
\begin{tabular}{ccc}
  \includegraphics[width=0.3\textwidth]{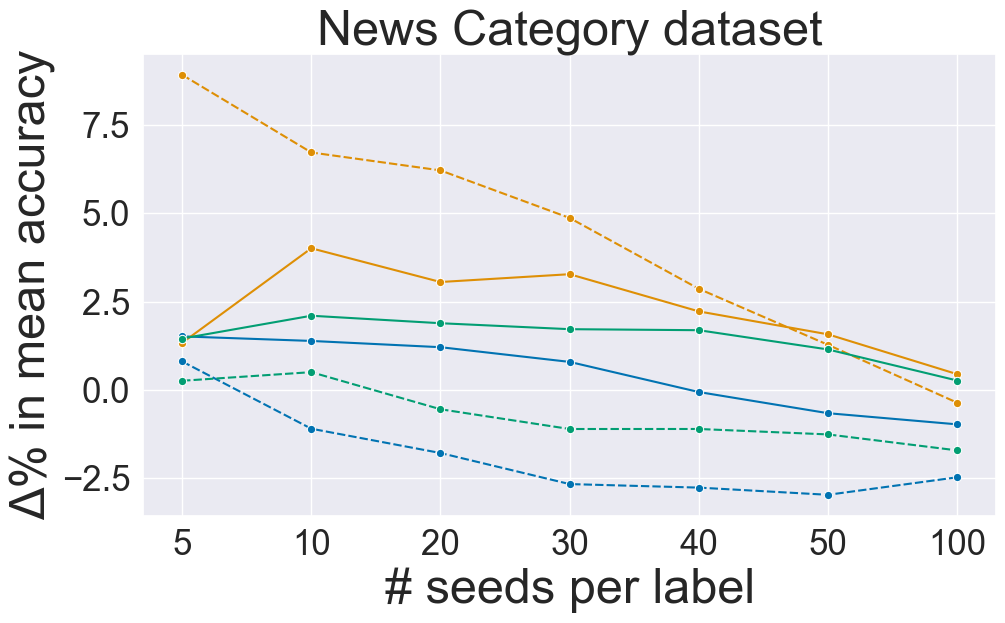} &   \includegraphics[width=0.3\textwidth]{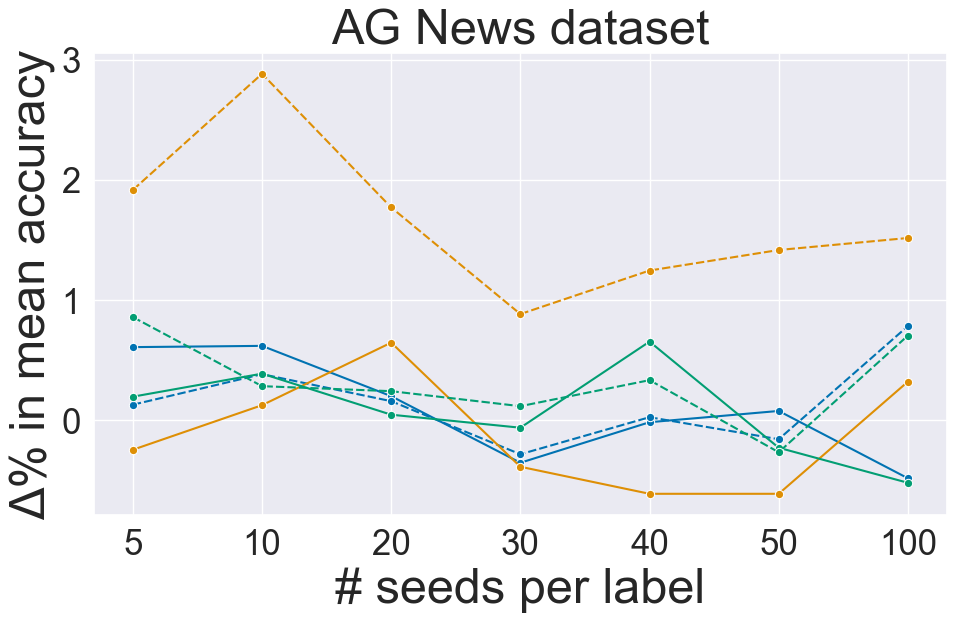}  & \includegraphics[width=0.3\textwidth]{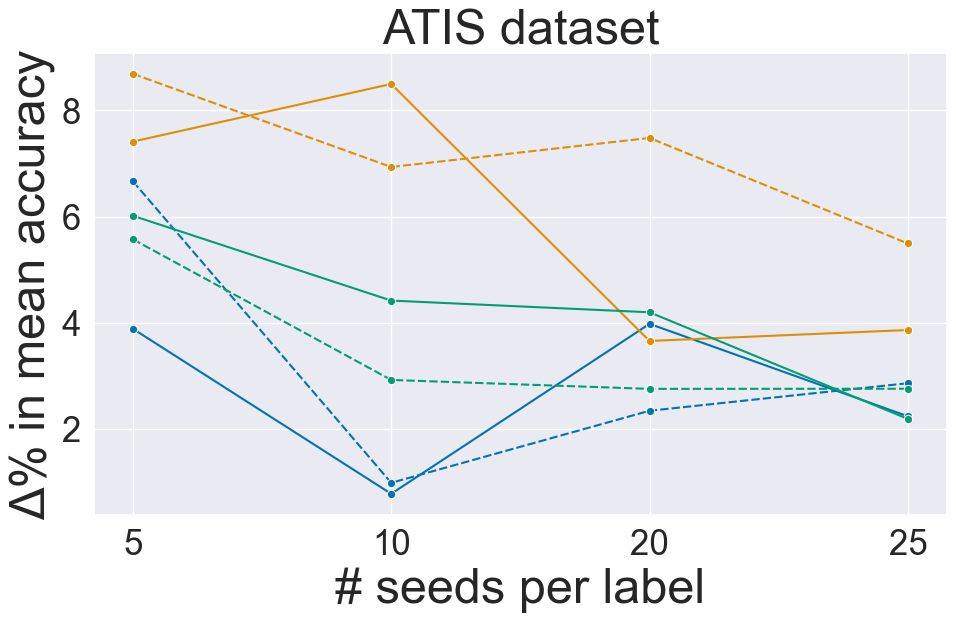} \\
  \includegraphics[width=0.3\textwidth]{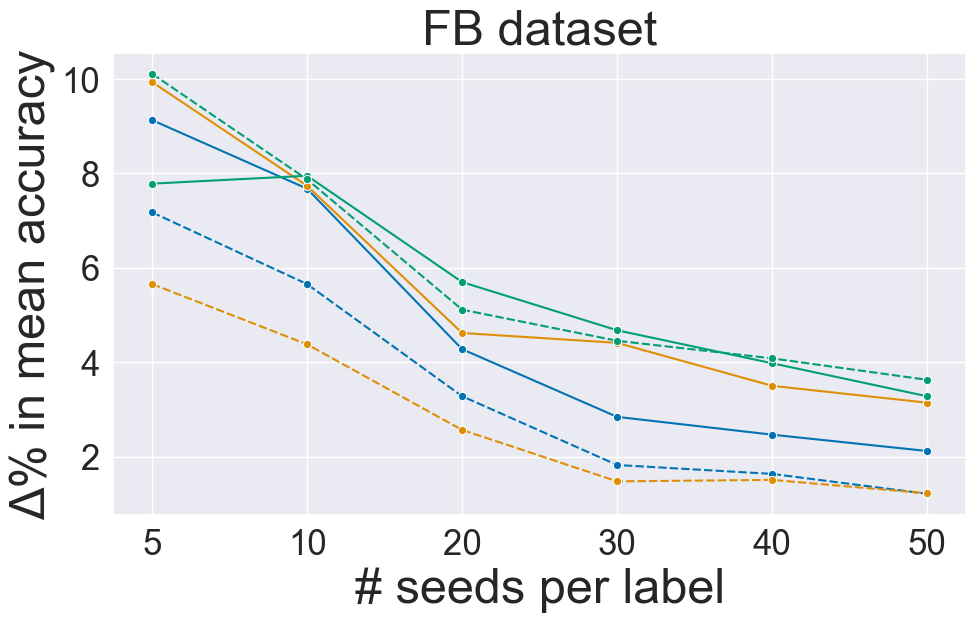} & \includegraphics[width=0.3\textwidth]{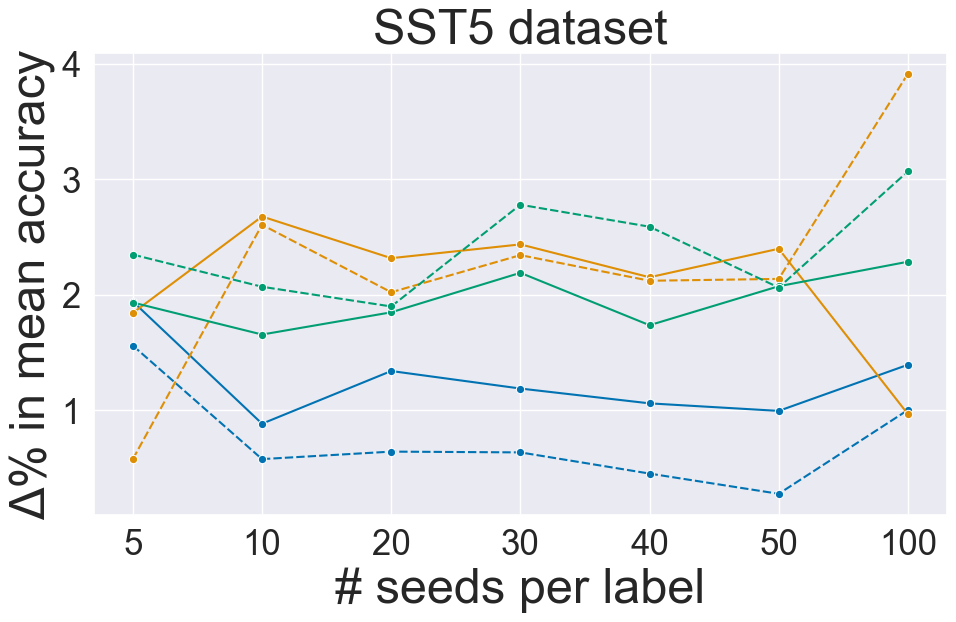} &   \includegraphics[width=0.3\textwidth]{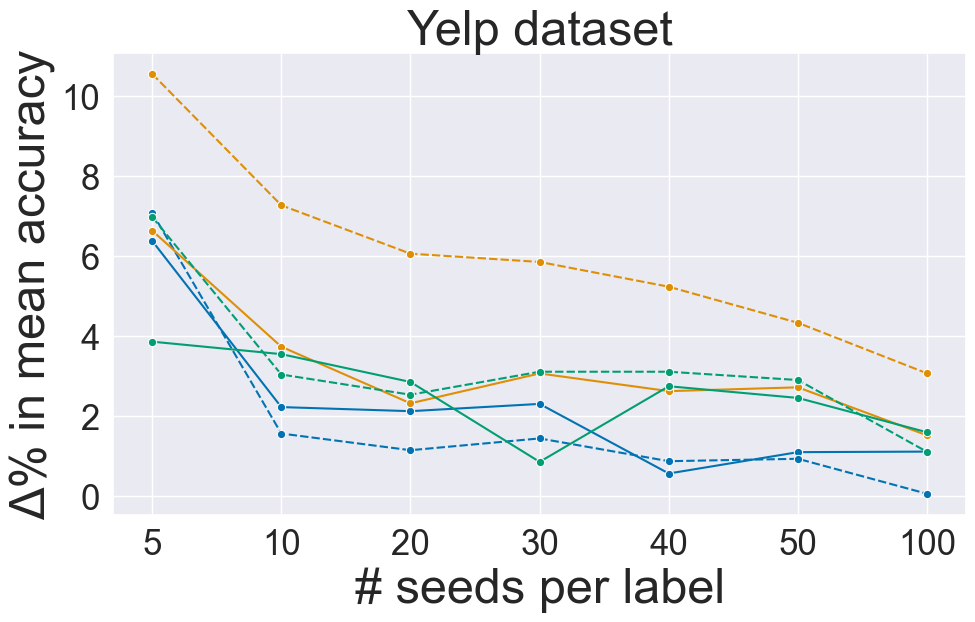} \\
  \multicolumn{3}{c}{\includegraphics[width=0.5\textwidth]{figures/legenda.png}}
\end{tabular}
\caption{The difference in mean accuracy for classifiers trained on the \emph{paraphrasing} augmentation method and the \emph{contextual swap} augmentation method for 6 different datasets. The \emph{paraphrasing} method works generally better in all cases with an decreasing  effect with increased number of seeds per label.}
\label{fig:mean_diff_perf_para_vs_swap}
\end{figure*}

We compared also other methods with the best performing LLM-based method (\emph{paraphrasing}) and the best established method (\emph{contextual insert} method). When comparing the \emph{paraphrasing} method and \emph{contextual swap} method, we can see based on Table~\ref{tab:count_perf_para_vs_swap} and Figure~\ref{fig:mean_diff_perf_para_vs_swap} that the \emph{paraphrasing method} is in nearly all cases better than the \emph{contextual swap} method, but the increased model accuracy decreases with number of seeds per label. A similar comparison can be seen for the \emph{paraphrasing} method and the \emph{backtranslation} method in Table~\ref{tab:count_perf_para_vs_back} and Figure~\ref{fig:mean_diff_perf_para_vs_back}.

\begin{figure*}[!t]
\begin{tabular}{ccc}
  \includegraphics[width=0.3\textwidth]{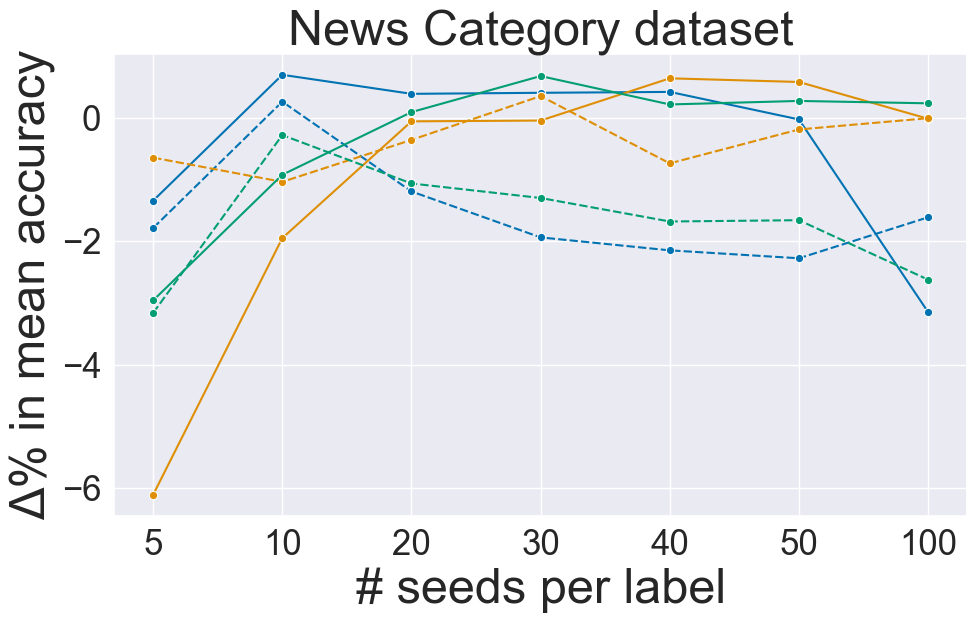} &   \includegraphics[width=0.3\textwidth]{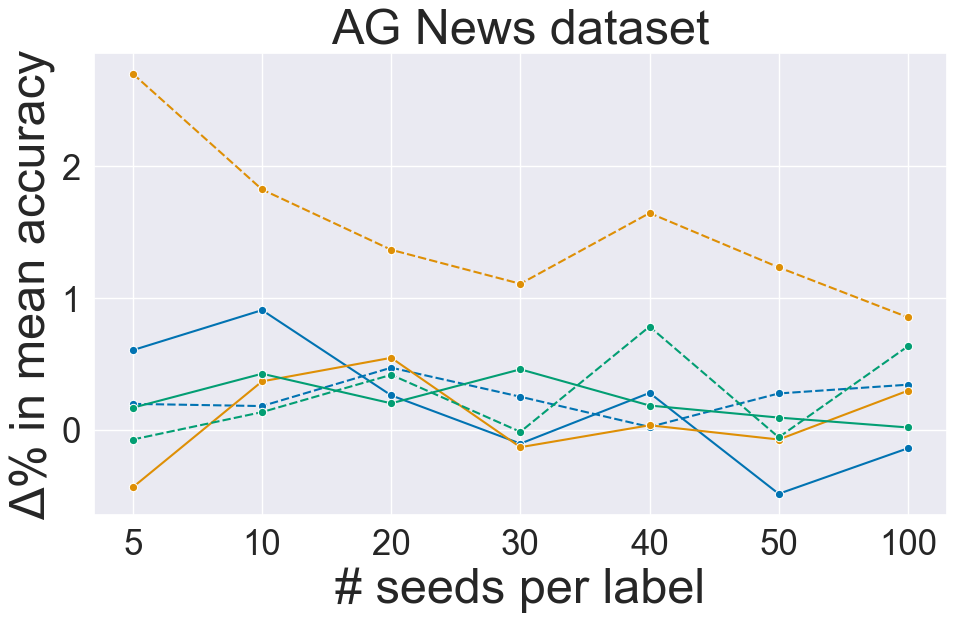}  & \includegraphics[width=0.3\textwidth]{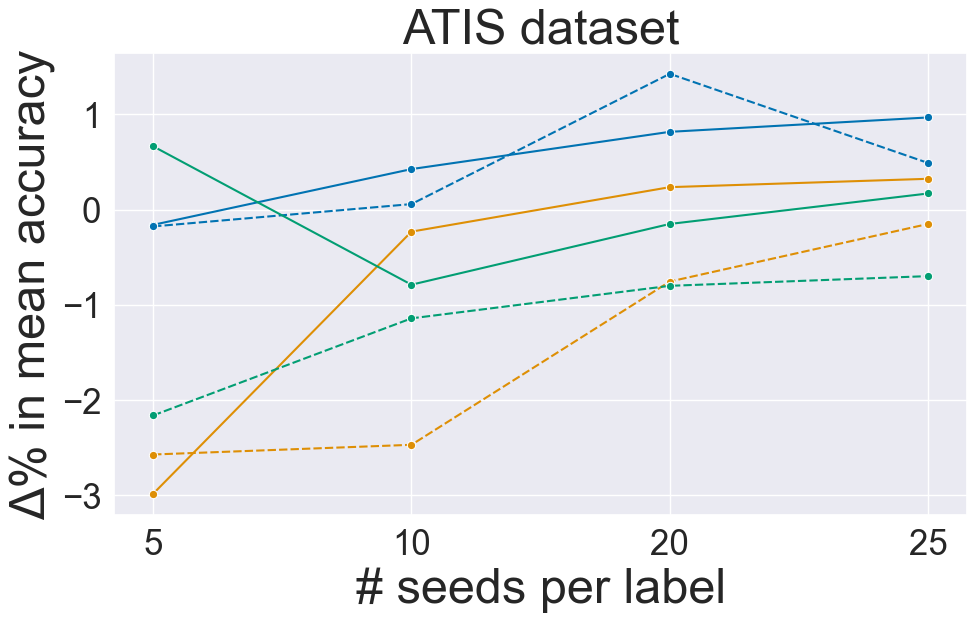} \\
  \includegraphics[width=0.3\textwidth]{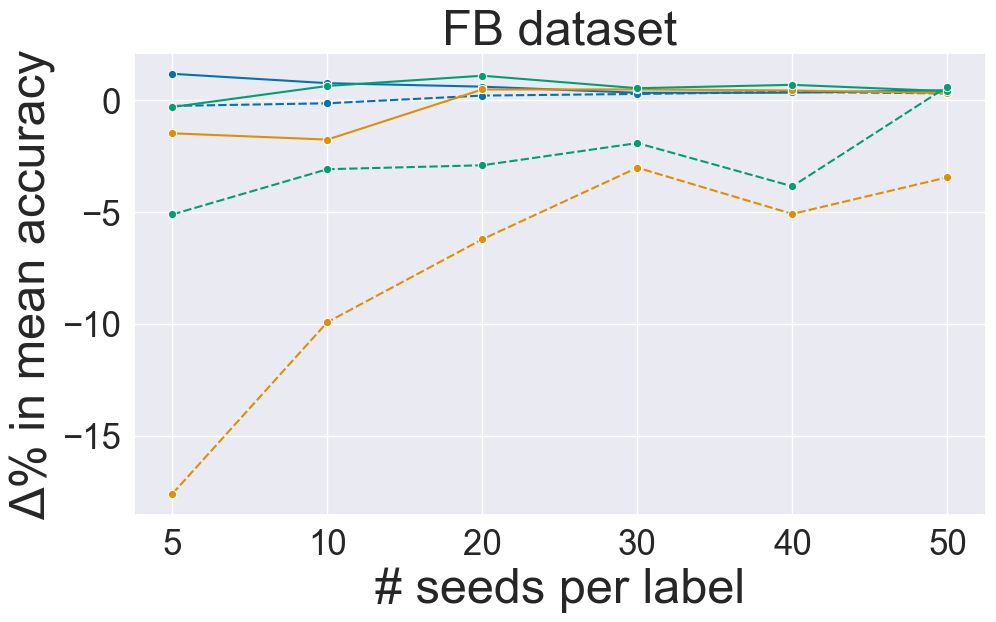} & \includegraphics[width=0.3\textwidth]{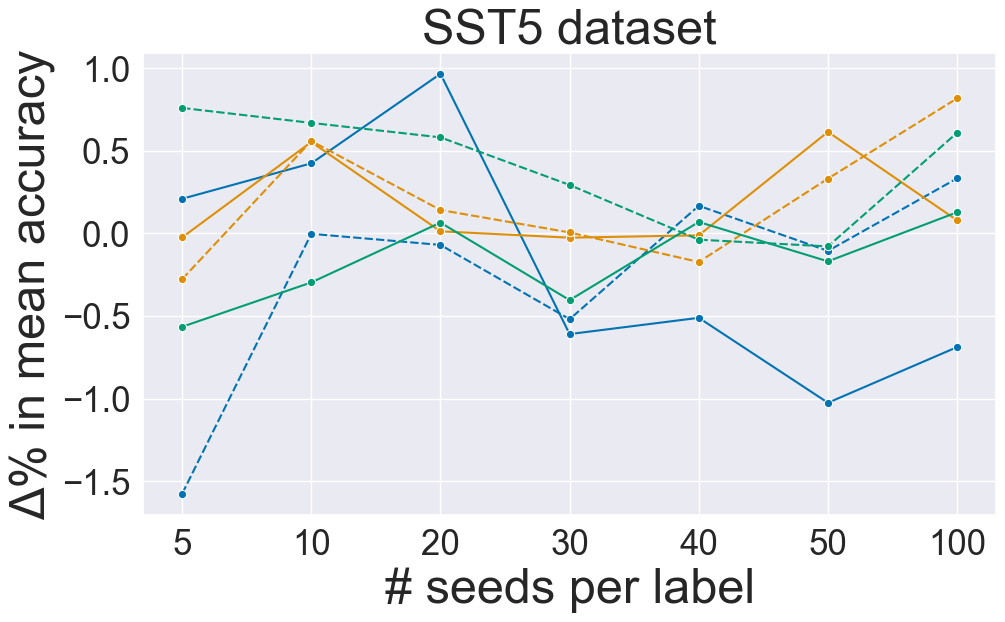} &   \includegraphics[width=0.3\textwidth]{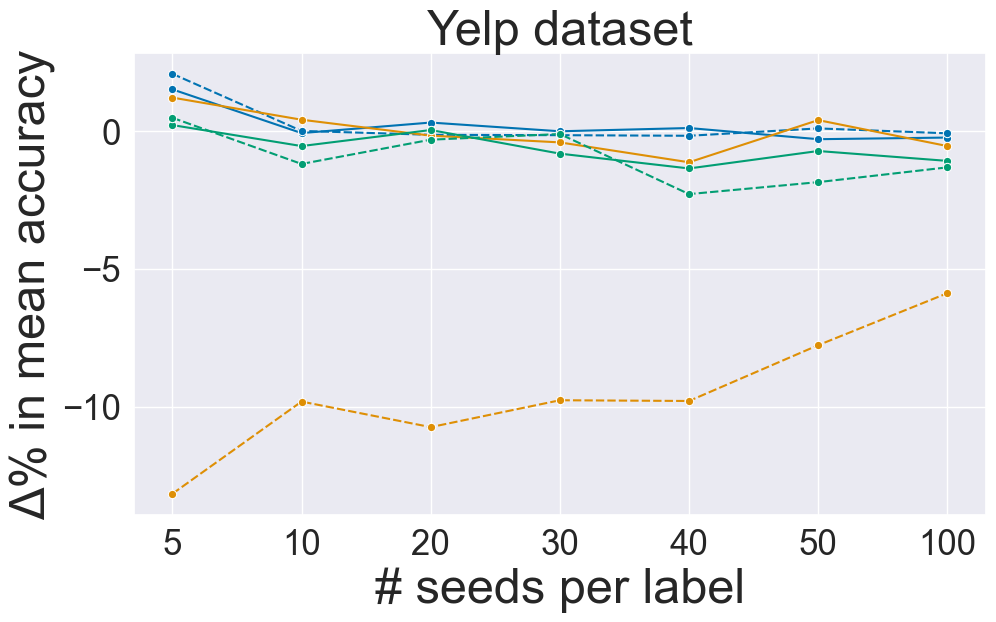} \\
  \multicolumn{3}{c}{\includegraphics[width=0.5\textwidth]{figures/legenda.png}}
\end{tabular}
\caption{The difference in mean accuracy for classifiers trained on the \emph{insert words} LLM-based augmentation method and the \emph{contextual insert} augmentation method for 6 different datasets. The cost of using the \emph{insert words} LLM-based method outweighs the benefits, as the \emph{contextual insert} method works in many cases slightly worse or outright better for model accuracy.}
\label{fig:mean_diff_perf_ins_vs_ins}
\end{figure*}

When comparing the best established method \emph{contextual insert} with the \emph{insert word} LLM-based method, we can see that it performs better for RoBERTa finetuning in Table~\ref{tab:count_perf_ins_vs_ins}, but the increase in model accuracy is not high as seen in Figure~\ref{fig:mean_diff_perf_ins_vs_ins}. When comparing the \emph{contextual insert} method and the \emph{swap word} LLM-based method the difference is even more in favour of the \emph{contextual insert} method as seen in Table~\ref{tab:count_perf_swap_vs_ins} and Figure~\ref{tab:count_perf_swap_vs_ins}. 

In general, the swap word methods performed the worst, while the insert words methods perfomed the best in cases of finetuning robustly pretrained models (RoBERTa) or for noisy datasets (SST-5, Yelp). 

\begin{table*}[t!]
    \centering
    \small
    \setlength\tabcolsep{5pt}
\begin{tabular}{@{}lc||c||c||c||c||c@{}}
\toprule
\textsc{Classifier}$\rightarrow$ & \multicolumn{2}{c}{\textsc{RoBERTa}} & \multicolumn{2}{c}{\textsc{BERT}} & \multicolumn{2}{c}{\textsc{DistilBERT}}\\
Dataset$\downarrow$  & Full & QLoRA & Full & QLoRA & Full & QLoRA\\
 \midrule
AG News & \texttt{~6 \textbf{\(|\)} 72 \textbf{\(|\)} ~0} & \texttt{17 \textbf{\(|\)} 44 \textbf{\(|\)} ~3} & \texttt{19 \textbf{\(|\)} 46 \textbf{\(|\)} ~0} & \texttt{42 \textbf{\(|\)} ~0 \textbf{\(|\)} ~0} & \texttt{~8 \textbf{\(|\)} 64 \textbf{\(|\)} ~2} & \texttt{37 \textbf{\(|\)} 10 \textbf{\(|\)} ~0} \\
 News Category & \texttt{~4 \textbf{\(|\)} 58 \textbf{\(|\)} ~9} & \texttt{~6 \textbf{\(|\)} ~4 \textbf{\(|\)} 34} & \texttt{27 \textbf{\(|\)} 30 \textbf{\(|\)} ~0} & \texttt{42 \textbf{\(|\)} ~0 \textbf{\(|\)} ~0} & \texttt{~7 \textbf{\(|\)} 60 \textbf{\(|\)} ~5} & \texttt{33 \textbf{\(|\)} ~6 \textbf{\(|\)} ~6} \\
ATIS & \texttt{14 \textbf{\(|\)} 16 \textbf{\(|\)} ~2} & \texttt{16 \textbf{\(|\)} 10 \textbf{\(|\)} ~3} & \texttt{16 \textbf{\(|\)} 16 \textbf{\(|\)} ~0} & \texttt{20 \textbf{\(|\)} ~8 \textbf{\(|\)} ~0} & \texttt{13 \textbf{\(|\)} 20 \textbf{\(|\)} ~1} & \texttt{20 \textbf{\(|\)} ~6 \textbf{\(|\)} ~1} \\
FB & \texttt{24 \textbf{\(|\)} 22 \textbf{\(|\)} ~1} & \texttt{36 \textbf{\(|\)} ~0 \textbf{\(|\)} ~0} & \texttt{32 \textbf{\(|\)} ~8 \textbf{\(|\)} ~0} & \texttt{36 \textbf{\(|\)} ~0 \textbf{\(|\)} ~0} & \texttt{28 \textbf{\(|\)} 16 \textbf{\(|\)} ~0} & \texttt{36 \textbf{\(|\)} ~0 \textbf{\(|\)} ~0} \\
SST-5 & \texttt{11 \textbf{\(|\)} 52 \textbf{\(|\)} ~5} & \texttt{37 \textbf{\(|\)} ~8 \textbf{\(|\)} ~1} & \texttt{22 \textbf{\(|\)} 38 \textbf{\(|\)} ~1} & \texttt{41 \textbf{\(|\)} ~2 \textbf{\(|\)} ~0} & \texttt{~4 \textbf{\(|\)} 76 \textbf{\(|\)} ~0} & \texttt{42 \textbf{\(|\)} ~0 \textbf{\(|\)} ~0} \\
YELP & \texttt{~4 \textbf{\(|\)} 52 \textbf{\(|\)} 12} & \texttt{16 \textbf{\(|\)} 32 \textbf{\(|\)} 10} & \texttt{22 \textbf{\(|\)} 36 \textbf{\(|\)} ~2} & \texttt{~6 \textbf{\(|\)} ~6 \textbf{\(|\)} 33} & \texttt{10 \textbf{\(|\)} 62 \textbf{\(|\)} ~1} & \texttt{37 \textbf{\(|\)} ~2 \textbf{\(|\)} ~4} \\
\bottomrule
    \end{tabular}
    \caption{Comparison of the number of cases where models trained using data from either \emph{paraphrasing} or \emph{backtranslation} method worked statistically (p=0.05) better or had similar accuracy when compared between each other. The numbers represent the result of one statistical test between 10 fine-tunings of the given model on data collected via the \emph{paraphrasing} or the \emph{backtranslation} using a specific random seed for a given number of seed samples per label. The cells are formatted in this way: \texttt{[\char"0023~paraphrasing was better]} \textbf{\(|\)} \texttt{[\char"0023~similar accuracy]} \textbf{\(|\)} \texttt{[\char"0023~backtranslation was better]}. In nearly all cases the \emph{paraphrasing} method works better for model accuracy, except for the Yelp dataset, where smaller changes from the \emph{backtranslation} might be more beneficial.}
    \label{tab:count_perf_para_vs_back}
\end{table*}

\begin{table*}[t!]
    \centering
    \small
    \setlength\tabcolsep{5pt}
\begin{tabular}{@{}lc||c||c||c||c||c@{}}
\toprule
\textsc{Classifier}$\rightarrow$ & \multicolumn{2}{c}{\textsc{RoBERTa}} & \multicolumn{2}{c}{\textsc{BERT}} & \multicolumn{2}{c}{\textsc{DistilBERT}}\\
Dataset$\downarrow$  & Full & QLoRA & Full & QLoRA & Full & QLoRA\\
 \midrule
AG News & \texttt{~4 \textbf{\(|\)} 70 \textbf{\(|\)} ~3} & \texttt{13 \textbf{\(|\)} 42 \textbf{\(|\)} ~8} & \texttt{~5 \textbf{\(|\)} 54 \textbf{\(|\)} 10} & \texttt{22 \textbf{\(|\)} 38 \textbf{\(|\)} ~1} & \texttt{~9 \textbf{\(|\)} 52 \textbf{\(|\)} ~7} & \texttt{17 \textbf{\(|\)} 36 \textbf{\(|\)} ~7} \\
News Category & \texttt{11 \textbf{\(|\)} 52 \textbf{\(|\)} ~5} & \texttt{~5 \textbf{\(|\)} ~8 \textbf{\(|\)} 33} & \texttt{32 \textbf{\(|\)} 18 \textbf{\(|\)} ~1} & \texttt{33 \textbf{\(|\)} 14 \textbf{\(|\)} ~2} & \texttt{29 \textbf{\(|\)} 26 \textbf{\(|\)} ~0} & \texttt{~5 \textbf{\(|\)} 22 \textbf{\(|\)} 26} \\
ATIS & \texttt{15 \textbf{\(|\)} 14 \textbf{\(|\)} ~2} & \texttt{15 \textbf{\(|\)} 14 \textbf{\(|\)} ~2} & \texttt{22 \textbf{\(|\)} ~4 \textbf{\(|\)} ~0} & \texttt{20 \textbf{\(|\)} ~8 \textbf{\(|\)} ~0} & \texttt{18 \textbf{\(|\)} 12 \textbf{\(|\)} ~0} & \texttt{20 \textbf{\(|\)} ~6 \textbf{\(|\)} ~1} \\
FB & \texttt{36 \textbf{\(|\)} ~0 \textbf{\(|\)} ~0} & \texttt{36 \textbf{\(|\)} ~0 \textbf{\(|\)} ~0} & \texttt{35 \textbf{\(|\)} ~2 \textbf{\(|\)} ~0} & \texttt{33 \textbf{\(|\)} ~6 \textbf{\(|\)} ~0} & \texttt{36 \textbf{\(|\)} ~0 \textbf{\(|\)} ~0} & \texttt{36 \textbf{\(|\)} ~0 \textbf{\(|\)} ~0} \\
SST-5 & \texttt{19 \textbf{\(|\)} 44 \textbf{\(|\)} ~1} & \texttt{22 \textbf{\(|\)} 26 \textbf{\(|\)} ~7} & \texttt{29 \textbf{\(|\)} 26 \textbf{\(|\)} ~0} & \texttt{31 \textbf{\(|\)} 22 \textbf{\(|\)} ~0} & \texttt{32 \textbf{\(|\)} 20 \textbf{\(|\)} ~0} & \texttt{39 \textbf{\(|\)} ~6 \textbf{\(|\)} ~0} \\
YELP & \texttt{23 \textbf{\(|\)} 38 \textbf{\(|\)} ~0} & \texttt{26 \textbf{\(|\)} 26 \textbf{\(|\)} ~3} & \texttt{33 \textbf{\(|\)} 18 \textbf{\(|\)} ~0} & \texttt{40 \textbf{\(|\)} ~4 \textbf{\(|\)} ~0} & \texttt{26 \textbf{\(|\)} 30 \textbf{\(|\)} ~1} & \texttt{37 \textbf{\(|\)} 10 \textbf{\(|\)} ~0} \\
\bottomrule
    \end{tabular}
    \caption{Comparison of the number of cases where models trained using data from either \emph{paraphrasing} or \emph{contextual swap} method worked statistically (p=0.05) better or had similar accuracy when compared between each other. The numbers represent the result of one statistical test between 10 fine-tunings of the given model on data collected via the \emph{paraphrasing} or the \emph{contextual swap} using a specific random seed for a given number of seed samples per label. The cells are formatted in this way: \texttt{[\char"0023~paraphrasing was better]} \textbf{\(|\)} \texttt{[\char"0023~similar accuracy]} \textbf{\(|\)} \texttt{[\char"0023~contextual swap was better]}. In most cases the \emph{paraphrasing} method works better for model accuracy in nearly all cases, with a higher accuracy of the \emph{contextual swap} method on news classification datasets.}
    \label{tab:count_perf_para_vs_swap}
\end{table*}

\begin{table*}[t!]
    \centering
    \small
    \setlength\tabcolsep{5pt}
\begin{tabular}{@{}lc||c||c||c||c||c@{}}
\toprule
\textsc{Classifier}$\rightarrow$ & \multicolumn{2}{c}{\textsc{RoBERTa}} & \multicolumn{2}{c}{\textsc{BERT}} & \multicolumn{2}{c}{\textsc{DistilBERT}}\\
Dataset$\downarrow$  & Full & QLoRA & Full & QLoRA & Full & QLoRA\\
 \midrule
AG News & \texttt{~5 \textbf{\(|\)} 74 \textbf{\(|\)} ~0} & \texttt{13 \textbf{\(|\)} 54 \textbf{\(|\)} ~2} & \texttt{~5 \textbf{\(|\)} 72 \textbf{\(|\)} ~1} & \texttt{21 \textbf{\(|\)} 40 \textbf{\(|\)} ~1} & \texttt{~6 \textbf{\(|\)} 68 \textbf{\(|\)} ~2} & \texttt{16 \textbf{\(|\)} 36 \textbf{\(|\)} ~8} \\
News Category & \texttt{~8 \textbf{\(|\)} 52 \textbf{\(|\)} ~8} & \texttt{~1 \textbf{\(|\)} 12 \textbf{\(|\)} 35} & \texttt{~4 \textbf{\(|\)} 58 \textbf{\(|\)} ~9} & \texttt{~3 \textbf{\(|\)} 54 \textbf{\(|\)} 12} & \texttt{~6 \textbf{\(|\)} 64 \textbf{\(|\)} ~4} & \texttt{~1 \textbf{\(|\)} ~6 \textbf{\(|\)} 38} \\
ATIS & \texttt{~9 \textbf{\(|\)} 28 \textbf{\(|\)} ~1} & \texttt{10 \textbf{\(|\)} 20 \textbf{\(|\)} ~4} & \texttt{~2 \textbf{\(|\)} 38 \textbf{\(|\)} ~3} & \texttt{~4 \textbf{\(|\)} 24 \textbf{\(|\)} ~8} & \texttt{~4 \textbf{\(|\)} 34 \textbf{\(|\)} ~3} & \texttt{~3 \textbf{\(|\)} ~2 \textbf{\(|\)} 20} \\
FB & \texttt{18 \textbf{\(|\)} 34 \textbf{\(|\)} ~1} & \texttt{20 \textbf{\(|\)} 18 \textbf{\(|\)} ~7} & \texttt{12 \textbf{\(|\)} 36 \textbf{\(|\)} ~6} & \texttt{12 \textbf{\(|\)} ~6 \textbf{\(|\)} 21} & \texttt{19 \textbf{\(|\)} 30 \textbf{\(|\)} ~2} & \texttt{26 \textbf{\(|\)} ~6 \textbf{\(|\)} ~7} \\
SST-5 & \texttt{~4 \textbf{\(|\)} 74 \textbf{\(|\)} ~1} & \texttt{~6 \textbf{\(|\)} 54 \textbf{\(|\)} ~9} & \texttt{~3 \textbf{\(|\)} 74 \textbf{\(|\)} ~2} & \texttt{~7 \textbf{\(|\)} 64 \textbf{\(|\)} ~3} & \texttt{~2 \textbf{\(|\)} 70 \textbf{\(|\)} ~5} & \texttt{15 \textbf{\(|\)} 38 \textbf{\(|\)} ~8} \\
YELP & \texttt{~4 \textbf{\(|\)} 74 \textbf{\(|\)} ~1} & \texttt{~5 \textbf{\(|\)} 64 \textbf{\(|\)} ~5} & \texttt{~6 \textbf{\(|\)} 62 \textbf{\(|\)} ~5} & \texttt{~0 \textbf{\(|\)} ~4 \textbf{\(|\)} 40} & \texttt{~0 \textbf{\(|\)} 76 \textbf{\(|\)} ~4} & \texttt{~5 \textbf{\(|\)} 30 \textbf{\(|\)} 22} \\
\bottomrule
    \end{tabular}
    \caption{Comparison of the number of cases where models trained using data from either \emph{insert words} LLM-based method or \emph{contextual insert} method worked statistically (p=0.05) better or had similar accuracy when compared between each other. The numbers represent the result of one statistical test between 10 fine-tunings of the given model on data collected via the \emph{insert words} or the \emph{contextual insert} using a specific random seed for a given number of seed samples per label. The cells are formatted in this way: \texttt{[\char"0023~insert words was better]} \textbf{\(|\)} \texttt{[\char"0023~similar accuracy]} \textbf{\(|\)} \texttt{[\char"0023~contextual insert was better]}. The \emph{swap word} method works well for RoBERTa full fine-tuning, but other than that the cases where it outperforms the \emph{contextual insert} method are equal to the cases where it is outperformed.}
    \label{tab:count_perf_ins_vs_ins}
\end{table*}

\begin{table*}[t!]
    \centering
    \small
    \setlength\tabcolsep{5pt}
\begin{tabular}{@{}lc||c||c||c||c||c@{}}
\toprule
\textsc{Classifier}$\rightarrow$ & \multicolumn{2}{c}{\textsc{RoBERTa}} & \multicolumn{2}{c}{\textsc{BERT}} & \multicolumn{2}{c}{\textsc{DistilBERT}}\\
Dataset$\downarrow$  & Full & QLoRA & Full & QLoRA & Full & QLoRA\\
 \midrule
AG News & \texttt{~1 \textbf{\(|\)} 80 \textbf{\(|\)} ~1} & \texttt{~9 \textbf{\(|\)} 52 \textbf{\(|\)} ~7} & \texttt{~4 \textbf{\(|\)} 76 \textbf{\(|\)} ~0} & \texttt{~0 \textbf{\(|\)} 70 \textbf{\(|\)} ~7} & \texttt{~7 \textbf{\(|\)} 68 \textbf{\(|\)} ~1} & \texttt{17 \textbf{\(|\)} 40 \textbf{\(|\)} ~5} \\
News Category & \texttt{~1 \textbf{\(|\)} 42 \textbf{\(|\)} 20} & \texttt{~0 \textbf{\(|\)} ~8 \textbf{\(|\)} 38} & \texttt{~0 \textbf{\(|\)} 46 \textbf{\(|\)} 19} & \texttt{~0 \textbf{\(|\)} 38 \textbf{\(|\)} 23} & \texttt{~0 \textbf{\(|\)} 54 \textbf{\(|\)} 15} & \texttt{~0 \textbf{\(|\)} ~2 \textbf{\(|\)} 41} \\
ATIS & \texttt{~0 \textbf{\(|\)} 20 \textbf{\(|\)} 14} & \texttt{~3 \textbf{\(|\)} 14 \textbf{\(|\)} 14} & \texttt{~0 \textbf{\(|\)} 12 \textbf{\(|\)} 18} & \texttt{~0 \textbf{\(|\)} 14 \textbf{\(|\)} 17} & \texttt{~1 \textbf{\(|\)} 14 \textbf{\(|\)} 16} & \texttt{~0 \textbf{\(|\)} ~2 \textbf{\(|\)} 23} \\
FB & \texttt{~0 \textbf{\(|\)} 18 \textbf{\(|\)} 27} & \texttt{~0 \textbf{\(|\)} ~4 \textbf{\(|\)} 34} & \texttt{~0 \textbf{\(|\)} 12 \textbf{\(|\)} 30} & \texttt{~0 \textbf{\(|\)} ~4 \textbf{\(|\)} 34} & \texttt{~0 \textbf{\(|\)} 20 \textbf{\(|\)} 26} & \texttt{~1 \textbf{\(|\)} ~0 \textbf{\(|\)} 35} \\
SST-5 & \texttt{~3 \textbf{\(|\)} 70 \textbf{\(|\)} ~4} & \texttt{~9 \textbf{\(|\)} 52 \textbf{\(|\)} ~7} & \texttt{~2 \textbf{\(|\)} 78 \textbf{\(|\)} ~1} & \texttt{~3 \textbf{\(|\)} 72 \textbf{\(|\)} ~3} & \texttt{~5 \textbf{\(|\)} 60 \textbf{\(|\)} ~7} & \texttt{15 \textbf{\(|\)} 32 \textbf{\(|\)} 11} \\
YELP & \texttt{~0 \textbf{\(|\)} 68 \textbf{\(|\)} ~8} & \texttt{~2 \textbf{\(|\)} 38 \textbf{\(|\)} 21} & \texttt{~1 \textbf{\(|\)} 54 \textbf{\(|\)} 14} & \texttt{~4 \textbf{\(|\)} 30 \textbf{\(|\)} 23} & \texttt{~0 \textbf{\(|\)} 54 \textbf{\(|\)} 15} & \texttt{~4 \textbf{\(|\)} 14 \textbf{\(|\)} 31} \\
\bottomrule
    \end{tabular}
    \caption{Comparison of the number of cases where models trained using data from either \emph{swap words} LLM-based method or \emph{contextual insert} method worked statistically (p=0.05) better or had similar accuracy when compared between each other. The numbers represent the result of one statistical test between 10 fine-tunings of the given model on data collected via the \emph{swap words} or the \emph{contextual insert} using a specific random seed for a given number of seed samples per label. The cells are formatted in this way: \texttt{[\char"0023~swap words was better]} \textbf{\(|\)} \texttt{[\char"0023~similar accuracy]} \textbf{\(|\)} \texttt{[\char"0023~contextual insert was better]}. The \emph{swap words} method is generally worse than the \emph{contextual insert} method for model accuracy.}
    \label{tab:count_perf_swap_vs_ins}
\end{table*}

\begin{figure*}[!t]
\begin{tabular}{ccc}
  \includegraphics[width=0.3\textwidth]{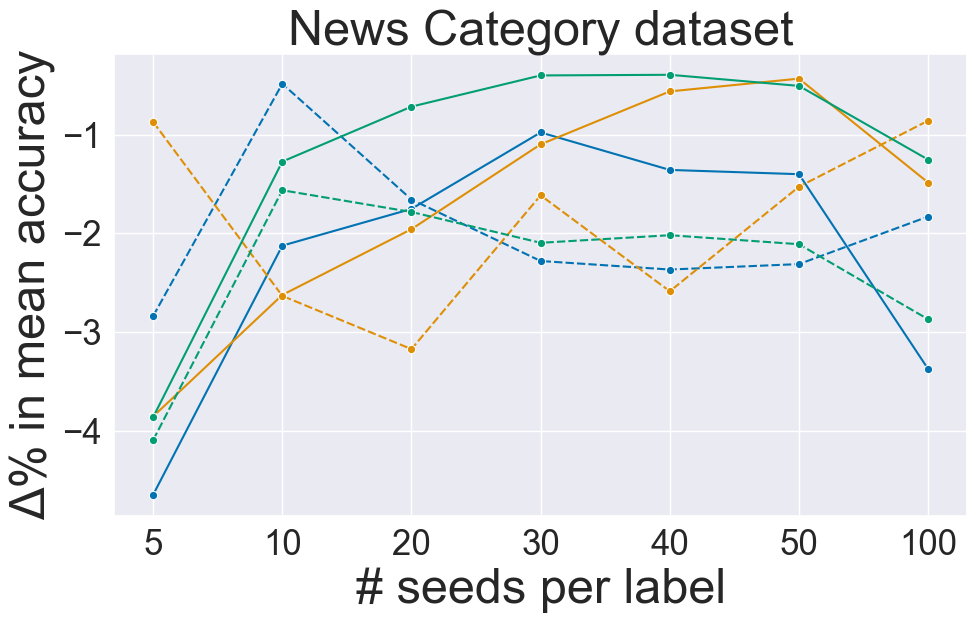} &   \includegraphics[width=0.3\textwidth]{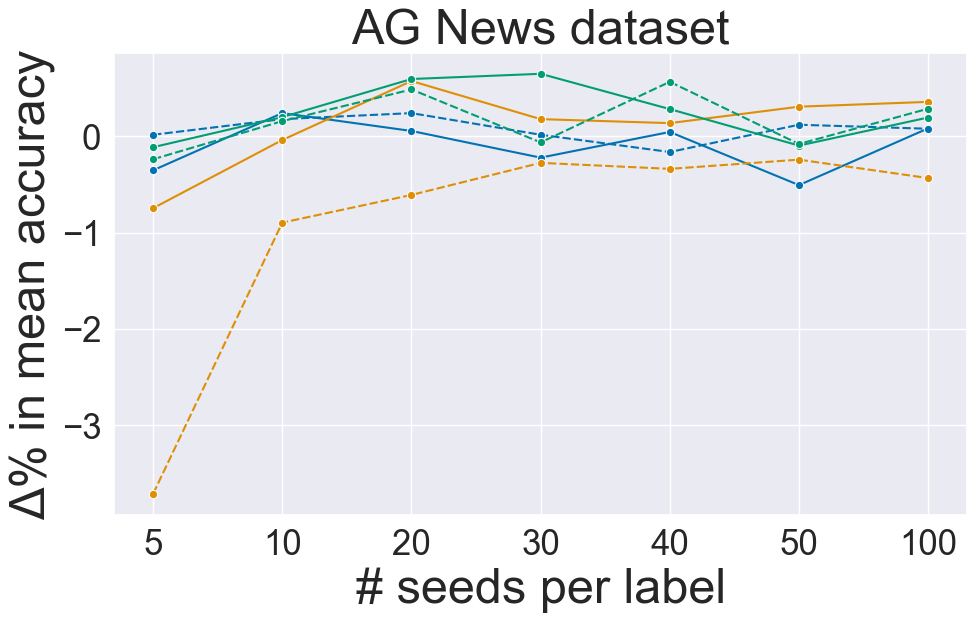}  & \includegraphics[width=0.3\textwidth]{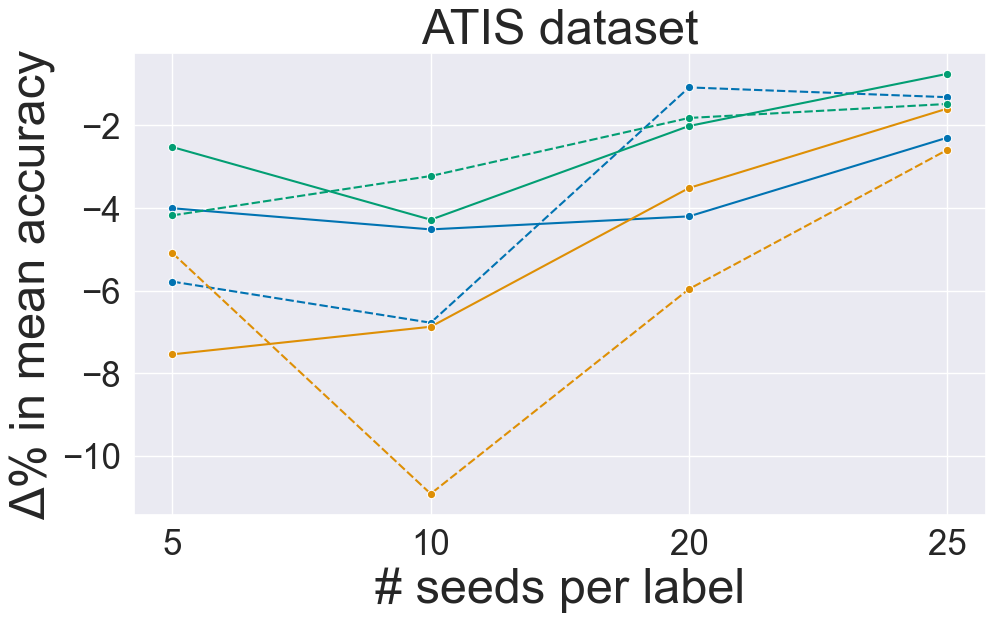} \\
  \includegraphics[width=0.3\textwidth]{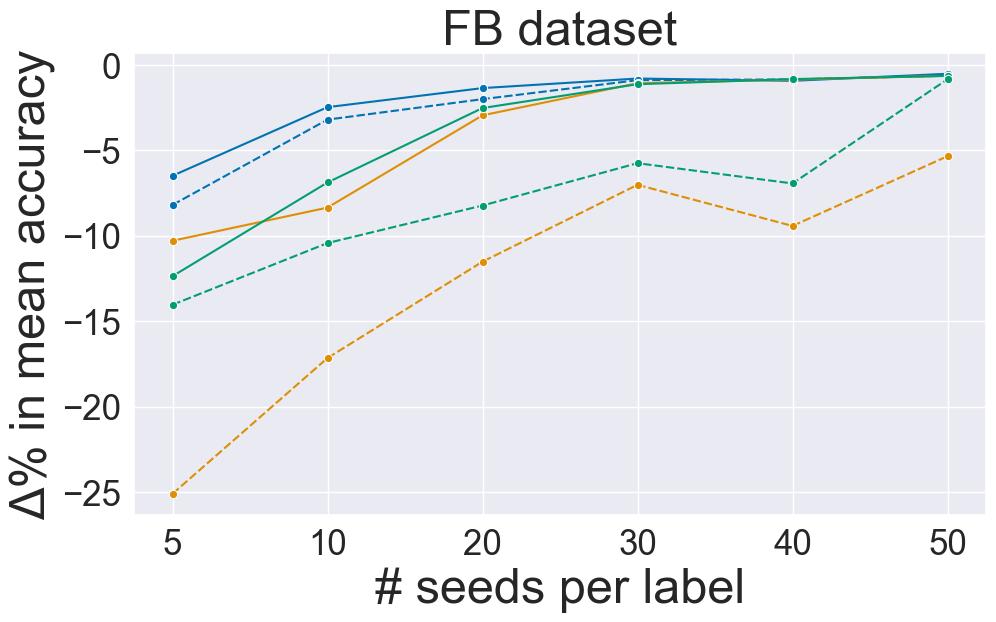} & \includegraphics[width=0.3\textwidth]{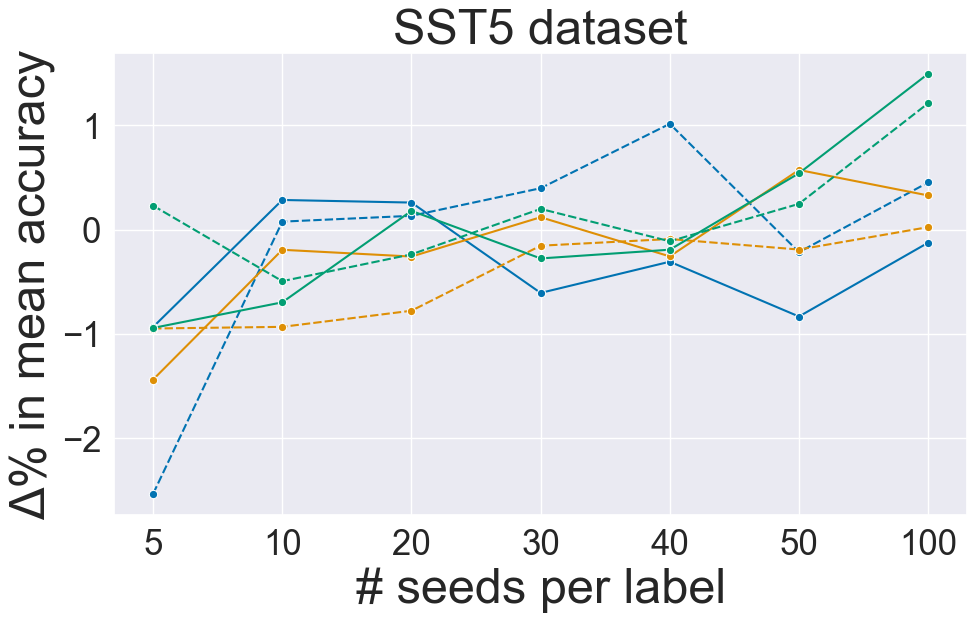} &   \includegraphics[width=0.3\textwidth]{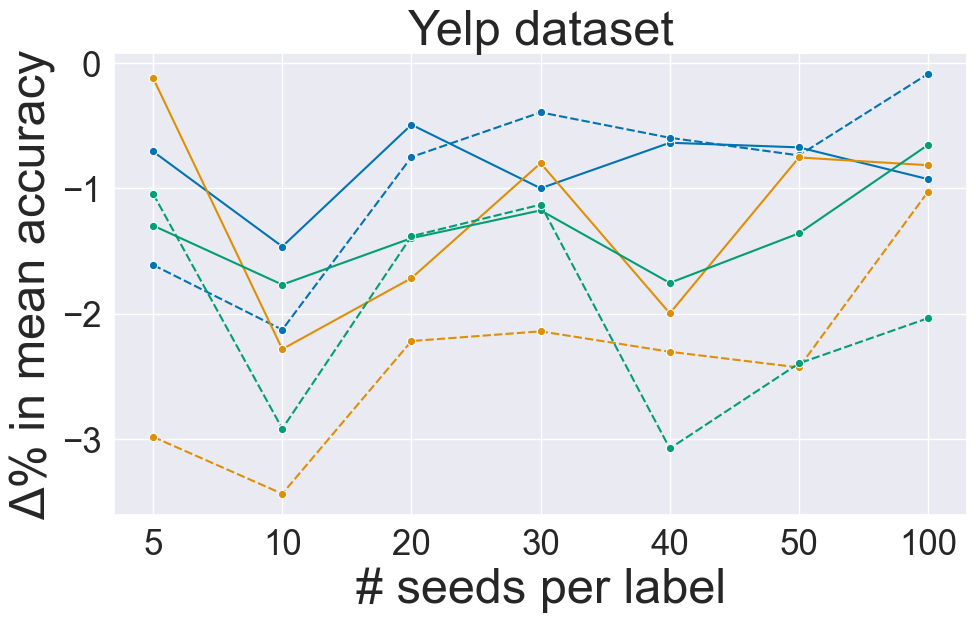} \\
  \multicolumn{3}{c}{\includegraphics[width=0.5\textwidth]{figures/legenda.png}}
\end{tabular}
\caption{The difference in mean accuracy for classifiers trained on the \emph{swap words} LLM-based augmentation method and the \emph{contextual insert} augmentation method for 6 different datasets. The cost of using the \emph{swap words} LLM-based method outweighs the benefits, as the \emph{contextual insert} method works in many cases  better for model accuracy.}
\label{fig:mean_diff_perf_ins_vs_swap}
\end{figure*}

\section{Comparison of augmentation methods increase for models accuracy against training only with seed samples}\label{sec:appendix_clean_only}

\begin{figure*}[!t]
\begin{tabular}{ccc}
  \includegraphics[width=0.3\textwidth]{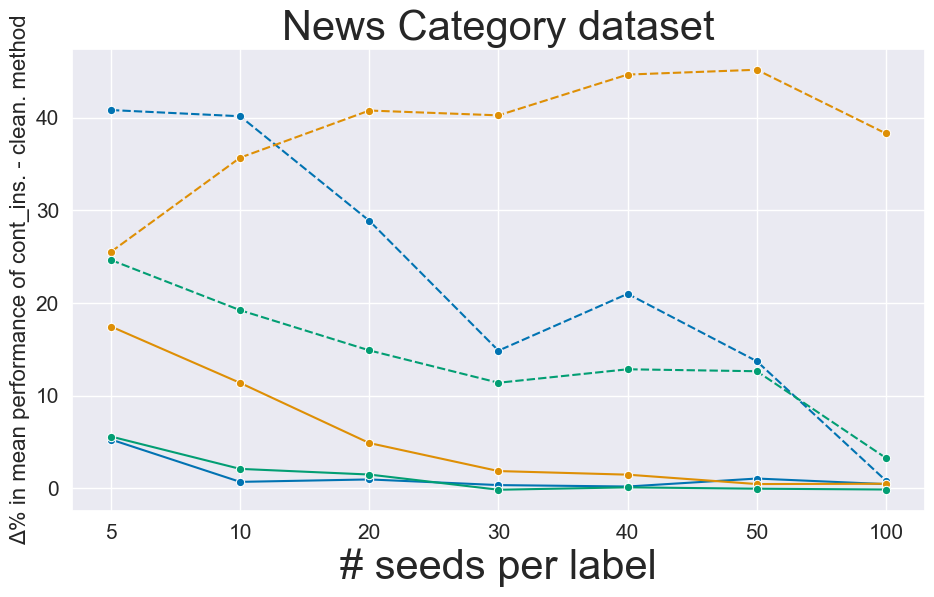} &   \includegraphics[width=0.3\textwidth]{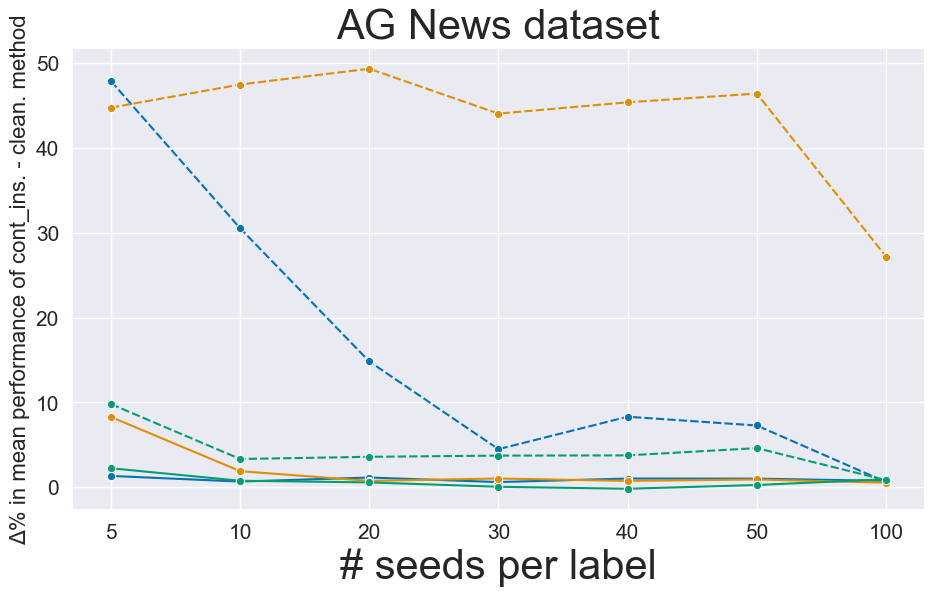}  & \includegraphics[width=0.3\textwidth]{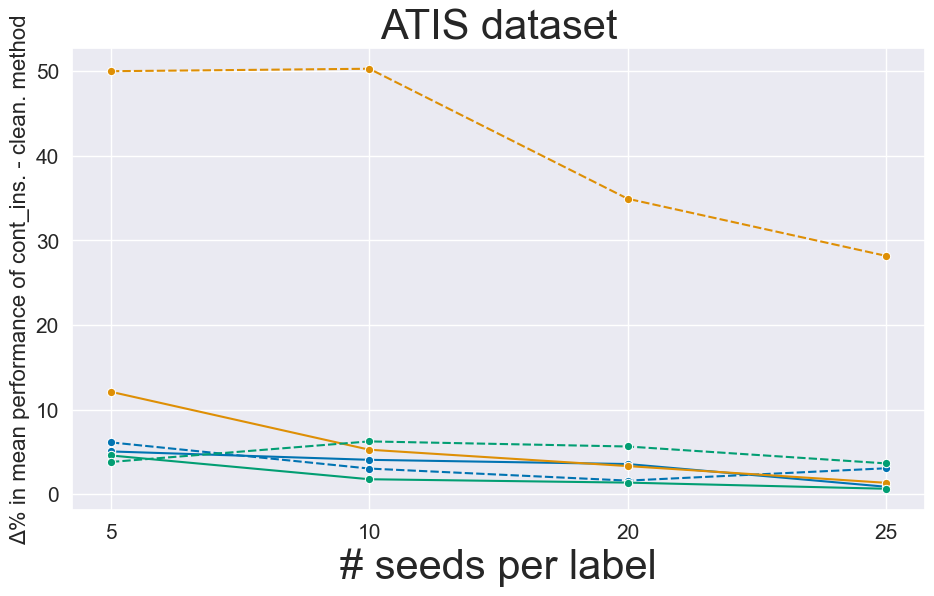} \\
  \includegraphics[width=0.3\textwidth]{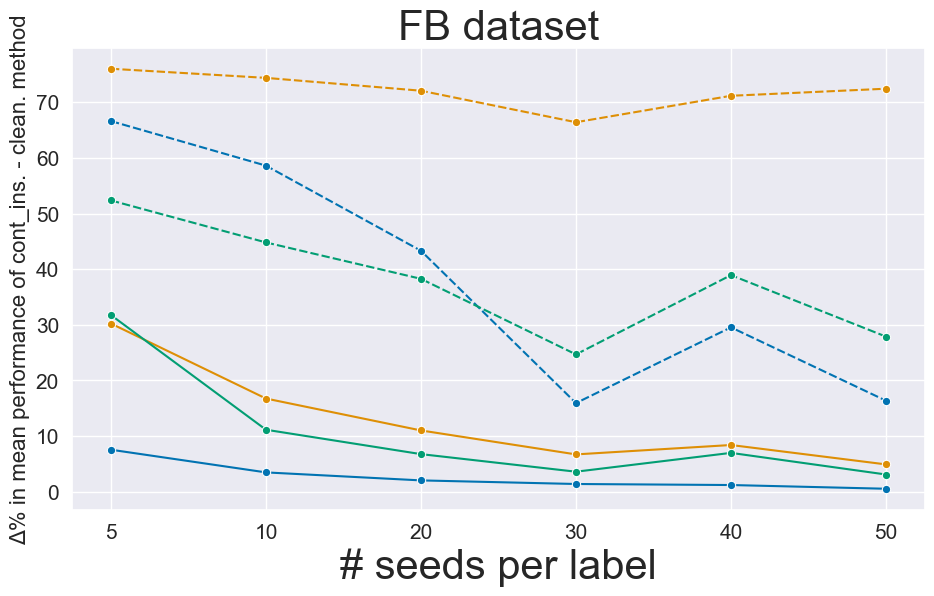} & \includegraphics[width=0.3\textwidth]{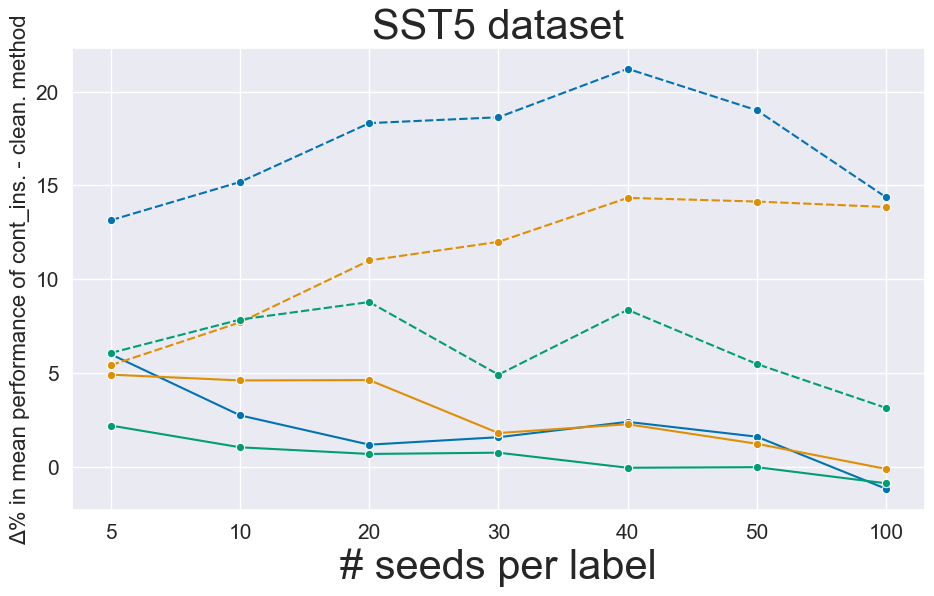} &   \includegraphics[width=0.3\textwidth]{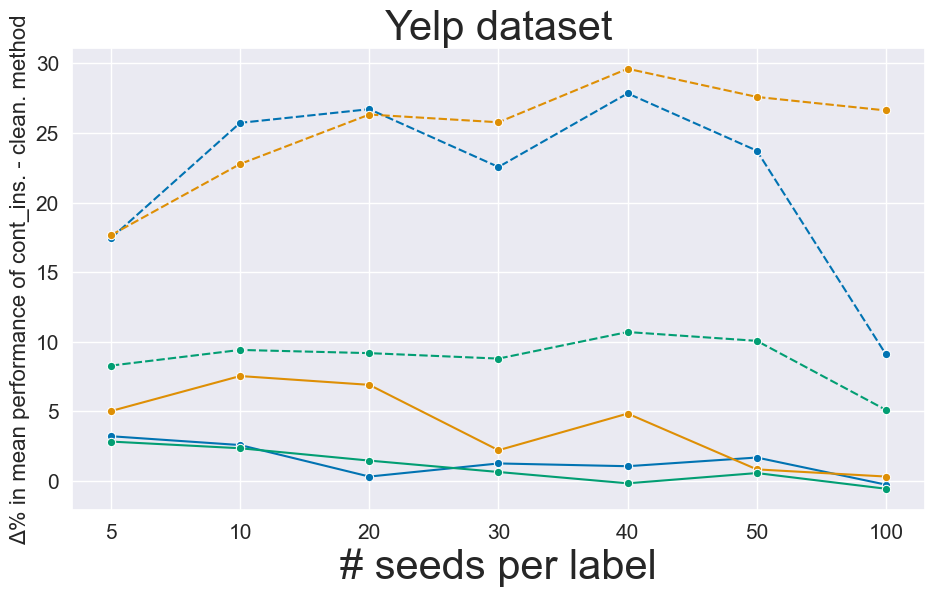} \\
  \multicolumn{3}{c}{\includegraphics[width=0.5\textwidth]{figures/legenda.png}}
\end{tabular}
\caption{The difference in mean accuracy for classifiers trained on the \emph{contextual insert} established augmentation method and using only the seed samples for fine-tuning for 6 different datasets. The cost of using the \emph{swap words} LLM-based method outweighs the benefits, as the \emph{accuracy insert} method works in many cases  better for model performance.}
\label{fig:mean_diff_perf_cont_ins_vs_clean}
\end{figure*}

\begin{figure*}[!t]
\begin{tabular}{ccc}
  \includegraphics[width=0.3\textwidth]{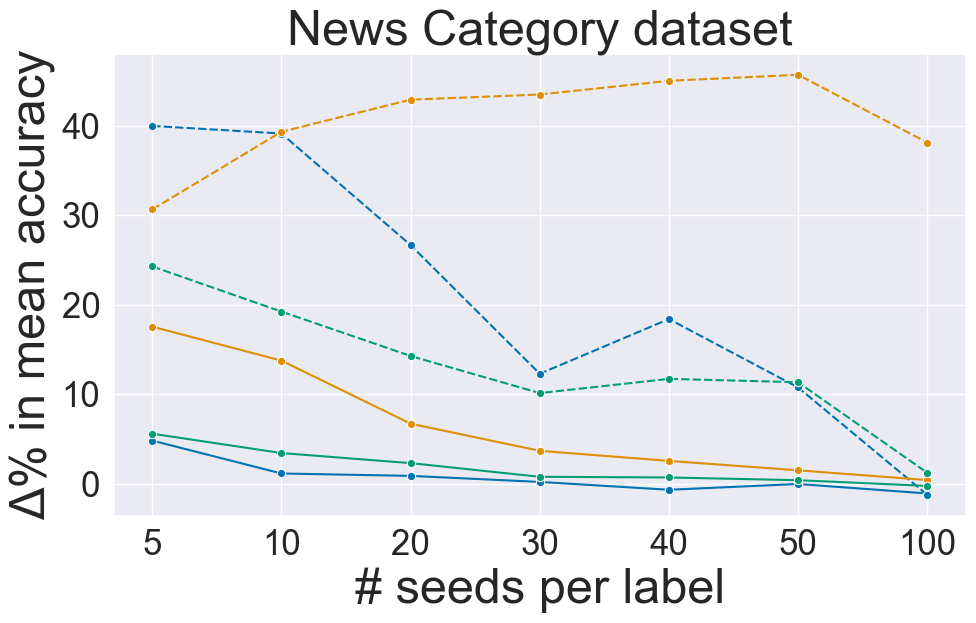} &   \includegraphics[width=0.3\textwidth]{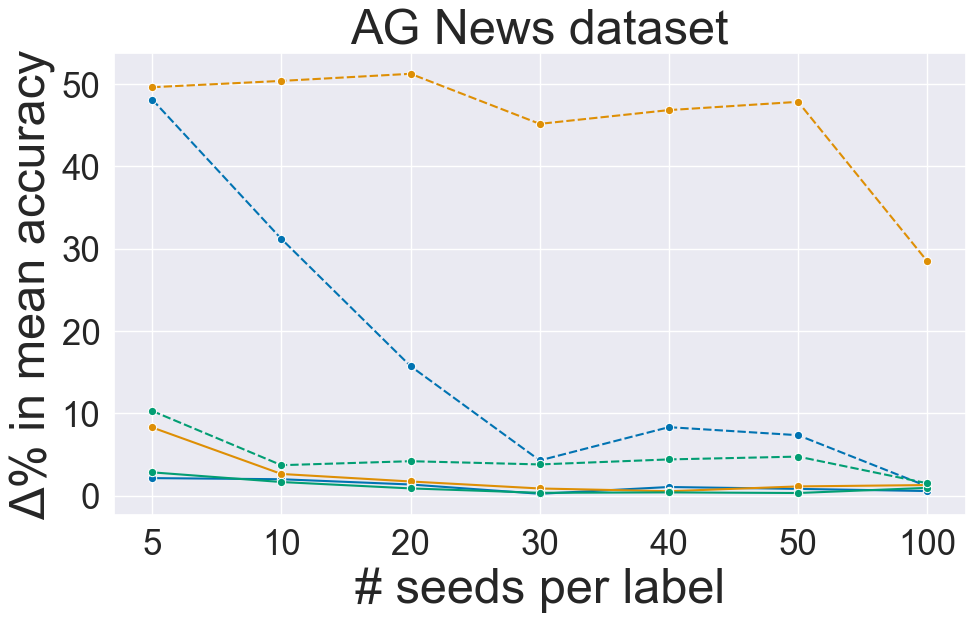}  & \includegraphics[width=0.3\textwidth]{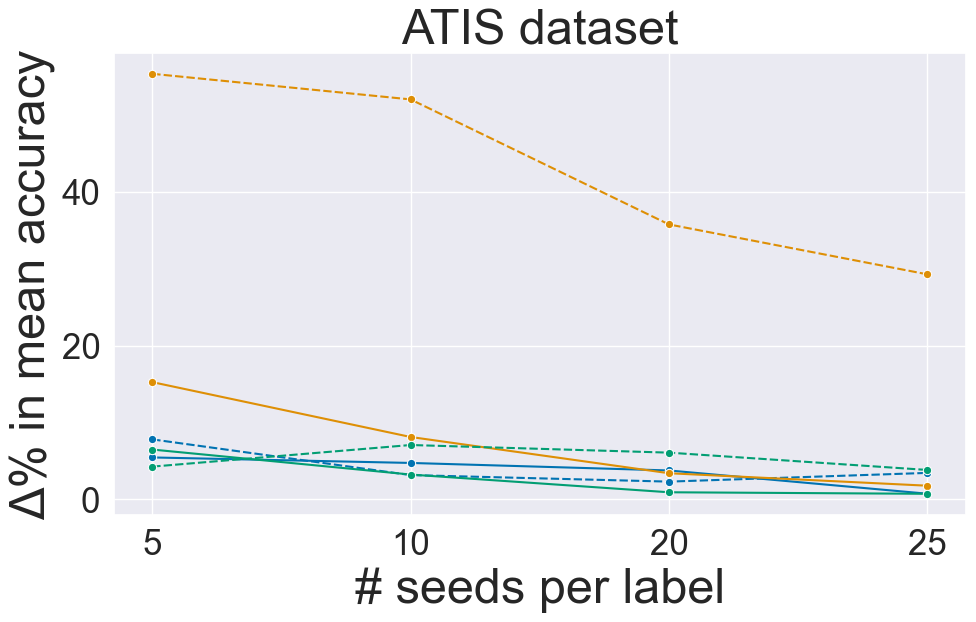} \\
  \includegraphics[width=0.3\textwidth]{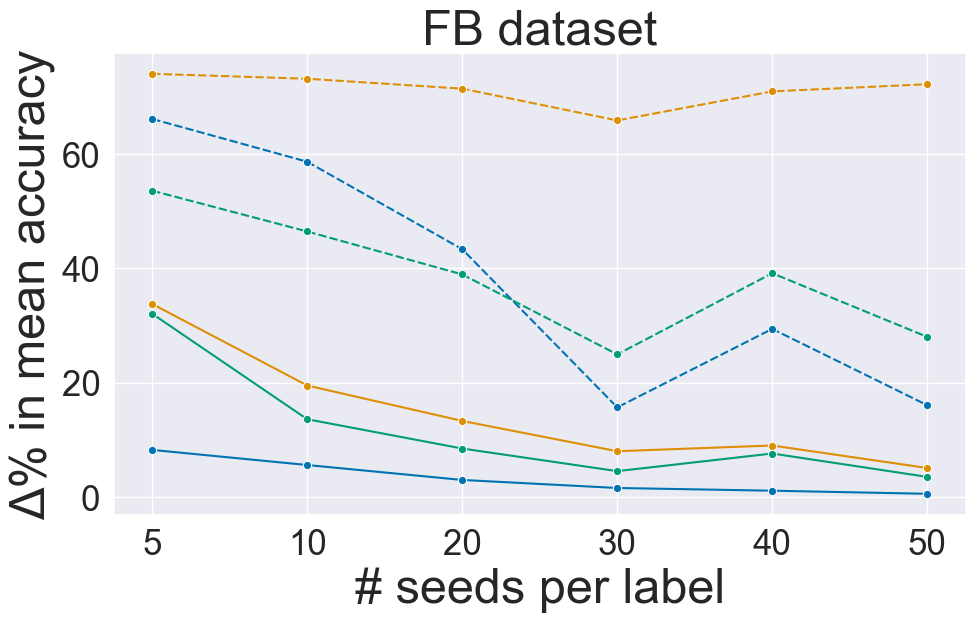} & \includegraphics[width=0.3\textwidth]{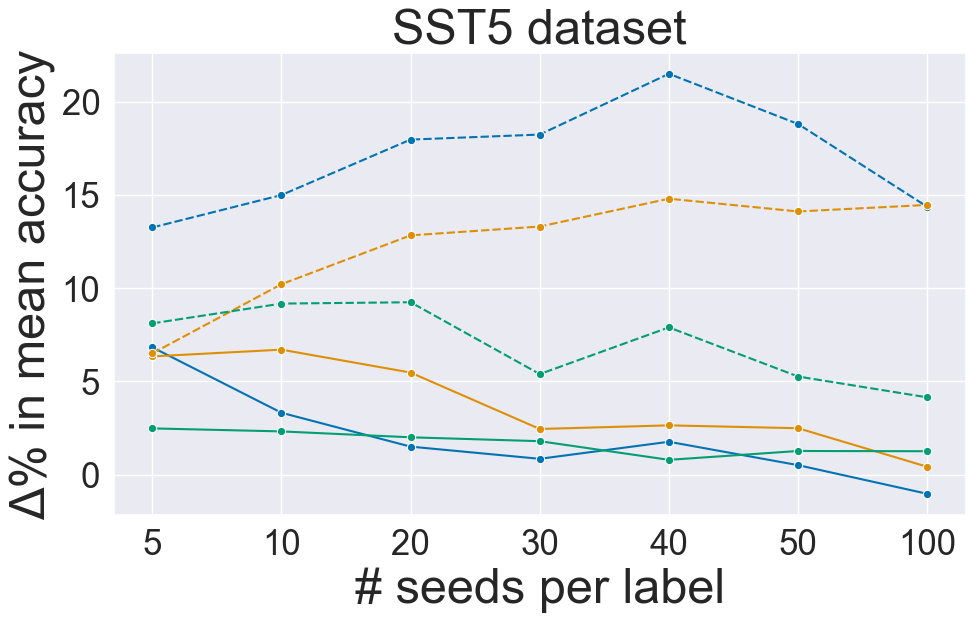} &   \includegraphics[width=0.3\textwidth]{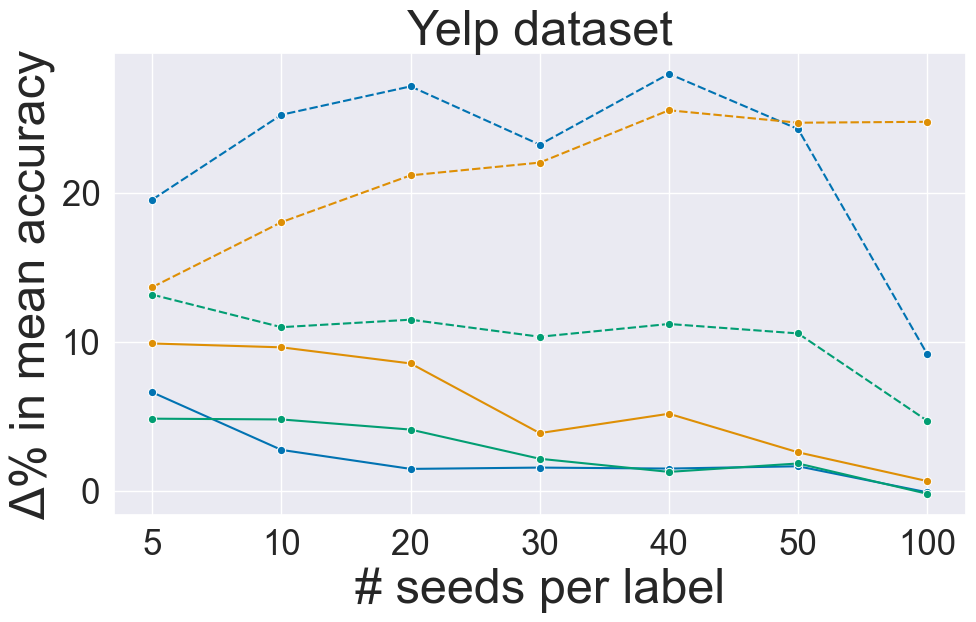} \\
  \multicolumn{3}{c}{\includegraphics[width=0.5\textwidth]{figures/legenda.png}}
\end{tabular}
\caption{The difference in mean accuracy for classifiers trained on the \emph{paraphrasing} LLM-based augmentation method and using only seed samples for fine-tuning for 6 different datasets. The cost of using the \emph{swap words} LLM-based method outweighs the benefits, as the \emph{contextual insert} method works in many cases  better for model accuracy.}
\label{fig:mean_diff_perf_para_vs_clean}
\end{figure*}

We compared the best LLM-based augmentation method \emph{paraphrasing} and the best established augmentation method \emph{contextual insert} and their effects on model accuracy when compared to models trained only using the seed samples. The results can be seen in Figures~\ref{fig:mean_diff_perf_cont_ins_vs_clean} and~\ref{fig:mean_diff_perf_para_vs_clean}. The LoRA fine-tuning methods have the highest relative and absolute increase for model accuracy even when considering increasing number of seed samples per label. Even though this increased accuracy decreases with number of seed samples used, this is most prominent for full fine-tuning, where cases of negative difference of mean accuracy exist. For LoRA finetuning the increased accuracy is still relatively high no matter the number of seeds per label. 

\section{Dataset details}\label{sec:appendix_dataset_details}

As we did not use all of the dataset labels and samples in each of the dataset, we list our setup here. We mostly used labels that were in the datasets with similar quantity to deal with the imbalanced datasets issue. All used datasets are in English language. For the \emph{News Category} dataset we used samples with labels \emph{politics}, \emph{wellness}, \emph{entertainment}, \emph{travel}, \emph{style and beauty} and \emph{parenting}. For the \emph{AG News}, \emph{SST-5} and \emph{Yelp} datasets we used all the samples. For the \emph{ATIS} dataset we used samples with labels \emph{atis\_abbreviation}, \emph{atis\_aircraft}, \emph{atis\_airfare} and \emph{atis\_flight\_time}. For the \emph{FB} dataset we used samples with labels \emph{get\_directions}, \emph{get\_distance}, \emph{get\_estimated\_arrival}, \emph{get\_estimated\_departure}, \emph{get\_estimated\_duration}, \emph{get\_info\_road\_condition} and \emph{get\_info\_traffic}. For the ATIS dataset we used values for number of seed samples per label [5, 10, 20, 25] and for the FB dataset we used values [5, 10, 20, 30, 40, 50] as both of these datasets had classes with fewer number of samples. 

\section{Established augmentation methods parameters used}\label{sec:appendix_established_params_used}

For the \emph{backtransaltion} method we used the \emph{facebook/wmt19-de-en} and \emph{facebook/wmt19-en-de models} models and set the maximum length of the produced translations to 300.

For the \emph{contextual insert} and \emph{contextual swap} methods same parameters were used: we considered 100 tokens for augmentation, with 30\% of the input text being changed with a minimum of 1 word and maximum of 10 words being either swapped or added and used \emph{BERT-large-uncased}~\footnote{https://huggingface.co/google-bert/bert-base-uncased} for our experiments. 

\section{LLM-based augmentation methods parameters and templates used}\label{sec:appendix_llm_params_used}

For GPT-3.5 data collection we used the \emph{gpt-3.5-turbo-0125} version of the model with \emph{temperature} of 1, \emph{top p} of 1 and \emph{presence penalty} at 0. For Llama3-8B we used the \emph{instruct} version~\footnote{https://huggingface.co/meta-Llama/Meta-Llama-3-8B-Instruct}, 4-bit quantization, max new tokens set at 1024, \emph{temperature} of 0.1 and \emph{top p} of 1. We collected 1 response for each seed sentence as we asked for 15 different augmentations in our prompts which are listed below. Both LLMs used the same prompts.

Paraphrasing prompt: \emph{Please provide 15 different changes of the Text by paraphrasing it. Output the full sentences. Output in format "1. sentence 1, 2. sentence 2, ... , 15. sentence 15". Text: "seed text placeholder".}

Insert words prompt: \emph{Please provide 15 different changes of the Text by inserting words into the Text. Output the full sentences. Output in format "1. sentence 1, 2. sentence 2, ... , 15. sentence 15". Text: "seed text placeholder".}

Swap words prompt: \emph{'Please provide 15 different changes of the Text by swapping words for their synonyms. Output the full sentences. Output in format "1. sentence 1, 2. sentence 2, ... , 15. sentence 15". Text: "seed text placeholder".}

\section{Classifier fine-tuning details}\label{sec:appendix_finetuning_details}

We selected the best hyperparameters after using hyperparameter search across models and classifiers. For both full-finetuning and LoRA finetuning, we used the same batch size across classifiers based on number of seed samples per label: we used 16 batch size for 5 to 20 seeds per label, 32 batch size for 20 to 30 seeds per label and 64 for 40 and more seeds per label. We used the same learning rate across classifiers set at \emph{1e-4}. We used AdamW optimizer in all cases.

For LoRA finetuning, we used \emph{r=16}, \emph{alpha=16}, \emph{dropout=0.1} and trained the model for 80 epochs. For full-finetuning, we performed the fine-tuning for 30 epochs.

\section{Best classifier model results}\label{sec:appendix_best_classifier_model}

\begin{figure*}[t!]
\begin{tabular}{cc}
  \includegraphics[width=0.475\textwidth]{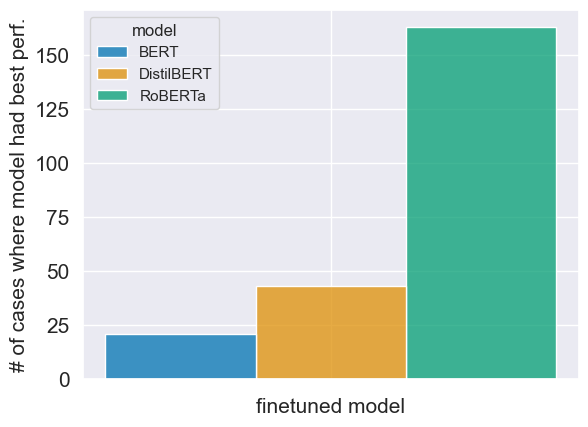} &   \includegraphics[width=0.475\textwidth]{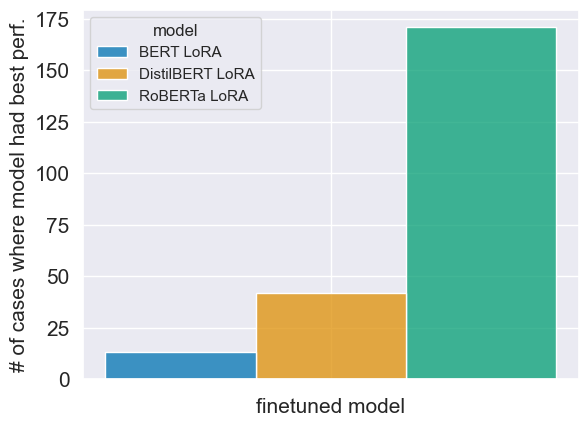} \\
\end{tabular}
\caption{No. cases where each model achieved highest accuracy for a particular combination of number of seeds, collected seeds, dataset when using full fine-tuning (left) and LoRA fine-tuning (right). These cases were gathered from the comparison of \emph{paraphrasing} LLM-based augmentation method and \emph{contextual insert} augmentation method.}
\label{fig:best_classifier_comparison}
\end{figure*}

We investigated which classifier performed best for both full fine-tuning and LoRA fine-tuning. We performed this analysis when comparing the \emph{paraphrasing} LLM-based method and \emph{contextual insert} method. We compared the cases with the same dataset, number of seed samples per label and random seeds used. In the majority of cases (approximately 80\% of the time) fine-tuned RoBERTa had the highest accuracy in all cases of fine-tuning, followed by DistilBERT and then BERT. The visualization of the results can be seen in Figure~\ref{fig:best_classifier_comparison}.

\end{document}